\documentclass{article}

\usepackage{PRIMEarxiv}

\usepackage[utf8]{inputenc}
\usepackage[T1]{fontenc}
\usepackage{url}
\usepackage{nicefrac}
\usepackage{microtype}
\usepackage{graphicx}
\usepackage{multirow}
\usepackage{amsmath,amssymb,amsfonts}
\usepackage{amsthm}
\usepackage{mathrsfs}
\usepackage[title]{appendix}
\usepackage{xcolor}
\usepackage{textcomp}
\usepackage{manyfoot}
\usepackage{booktabs}
\usepackage{subcaption}
\usepackage{algorithm}
\usepackage{algorithmicx}
\usepackage{algpseudocode}
\usepackage{listings}
\usepackage{lmodern,babel,adjustbox}
\usepackage{pgfplots}
\pgfplotsset{compat=newest}
\usepackage{tikz}
\usetikzlibrary{patterns}
\usepackage[numbers,sort&compress]{natbib}
\usepackage{hyperref}
\usepackage{cleveref}

\newenvironment{tablenotes}{\par\vspace{0.3em}\begin{list}{}{\setlength{\leftmargin}{1.5em}\setlength{\itemsep}{0pt}\setlength{\parsep}{0pt}}}{\end{list}}

\theoremstyle{plain}

\theoremstyle{definition}

\newcommand{\eg}{e.g.}

\newcommand{\sysname}{DumpsterCluster}

\graphicspath{{figures/}}

\renewcommand{\shorttitle}{}
\title{\raisebox{-1.18cm}{\includegraphics[width=3cm]{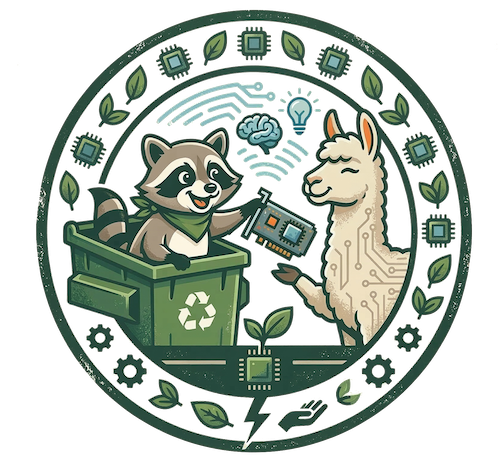}}DumpsterCluster: From Dumpster Diving to Serving LLaMA-70B on \$60 GPUs}

\author{
    Zeyu Cao\textsuperscript{1} \And
    Xuan Guo\textsuperscript{1,3} \And
    Cheng Zhang \AND
    Cheuk Hang Lau\textsuperscript{3} \And
    Ilia Shumailov\textsuperscript{2} \And
    Yiren Zhao\textsuperscript{1} \AND   
    \quad \textnormal{\textsuperscript{1}University of Cambridge 
    \quad
    \textsuperscript{2}University of Oxford
    \quad
    \textsuperscript{3}Quettaflop AI
}
}

\begin{document}

\maketitle

\begin{abstract}
As AI datacenters retire functional GPUs, vast quantities of still capable accelerators enter secondary markets. This paper investigates whether these ``retired'' GPUs can find a productive afterlife to form a \textit{\sysname{}} that can serve modern LLM inference, and under what conditions such repurposing is economically viable and environmentally sustainable.
We physically built a 128-GPU \sysname{} from scratch using only second-hand components and ran it for one year. At current market prices (\$22K for the \sysname{} vs. \$600K for an 8-GPU B200 system), the economic advantages are substantial. Through pipeline-parallel optimizations, our V100 based \sysname{}  achieves competitive LLaMA-70B throughput, validating production viability.
However, our deployment reveals critical context dependencies. Older GPUs consume significantly more energy per token, making total cost of ownership favorable only in regions with inexpensive electricity. Under grid-average carbon intensity, second-hand systems can produce approximately $4\times$ higher total carbon emissions per token for 8B models, and over $40\times$ for 70B models, compared to current-generation hardware.
These findings show that GPU afterlife is not universally sustainable -- hardware repurposing must be strategically coupled with low carbon energy sources. When deployed in regions with favourable energy economics and clean electricity, second-hand GPUs offer a viable pathway for expanding AI capacity while advancing affordability, energy security, and environmental responsibility.
\end{abstract}

\section{Introduction}\label{sec1}
The rapid rise of large-scale AI models has triggered unprecedented demand for computational infrastructure \cite{dubey2024llama,achiam2023gpt}. Hardware vendors have responded with accelerated GPU release cycles \cite{choquette2023nvidia,AMD}, each generation offering improved performance but requiring priority access to constrained semiconductor manufacturing capacity. This continuous upgrade cycle has created a peculiar phenomenon: datacenters regularly retire still-functional GPUs to make room for newer hardware, creating a growing secondary market of ``discarded'' but capable accelerators.

This pattern raises concerns about sustainability, supply-chain security, and accessibility. From a sustainability perspective, continual manufacture of new accelerators and premature retirement of functional hardware impose substantial embodied carbon costs from semiconductor fabrication while contributing to mounting electronic waste. From a supply-chain perspective, the concentration of advanced GPU production among few vendors and foundries has made access to frontier hardware increasingly constrained, with many organizations facing high prices, long procurement delays, and structural dependence on fragile supply chains. From an accessibility perspective, the escalating costs of current-generation systems -- a single 8-GPU B200 node costs \$600K -- place state-of-the-art AI infrastructure beyond reach for many researchers and organizations.
These pressures motivate a simple but underexplored alternative: giving retired GPUs a productive afterlife by repurposing them for modern AI workloads. Such devices are available on secondary markets at a fraction of new-hardware costs, yet remain computationally capable. This raises a fundamental question: Can second-hand GPUs form a \textit{\sysname{}} to support LLM inference in a way that is economically viable, resilient to supply constraints, and genuinely compatible with environmental and energy-security objectives?

To answer this, we designed and built a 128-GPU \sysname{} from the ground up using entirely second-hand components -- not only GPU accelerators, but also CPU processors, memory modules, and motherboards sourced from the secondary market. We deployed this cluster for one year, serving real LLM inference workloads.
The complete hardware configuration from the real build is presented in \Cref{fig:cluster-config}. Each rack-level node comprises pre-owned motherboards, CPUs, RAM, and NVIDIA V100 GPUs. On the software side, we developed a custom LLM serving engine in Rust implementing a device-level pipeline parallelism strategy not found in commodity serving systems. We empirically demonstrate that this serving approach enables older GPUs, despite their restricted HBM capacity, bandwidth, and GPU-to-GPU interconnect speeds, to effectively execute inference for state-of-the-art LLMs. Tokens Per Second (TPS) scales linearly with device count, enabling cost-effective inference compared to modern GPUs such as H100s.
In this paper, we present the first in-depth analysis of the economic and environmental trade-offs of repurposing retired hardware for LLM inference, revealing critical dependencies on electricity cost and carbon intensity that determine when second-hand GPU deployment is truly economically and environmentally sustainable.

\section{Method}\label{sec:method}

\subsection{The advantages and drawbacks of second-hand devices}

We mostly consider second-hand hardware devices that are batch-deployed and retired from commercial data centers after they are superseded by a new generation of hardware. In 2023 alone, NVIDIA shipped more than 3 million GPU chips to data centers~\cite{eadline_2024}. Many of these data center GPUs are scheduled to be phased out approximately every three years to make way for new product release cycles, resulting in millions of second-hand GPUs entering the market with each cycle.

In this study, we focus primarily on NVIDIA V100 GPUs, which are two generations behind the widely deployed Hopper series (\eg{} H100 and H200) and three generations older than the Blackwell (\eg{} B100 and B200).
V100 GPUs are now widely available on the second-hand market after multiple waves of decommission~\cite{jade_hpc_2024, microsoft_2025, microsoft_azure_2025, procurri_2024}.
The primary advantages of using such devices are their cost-effectiveness and reduced amortized embodied carbon. In addition, as LLM model serving becomes a major workload, reusing second-hand hardware also contributes to circular economy and addresses the e-waste concern~\cite{wang2024waste}.
However, there are also significant drawbacks of pre-owned devices, including their potentially high operational carbon and costs from naive deployments, as well as concerns regarding their reliability.

\noindent \textbf{Price competitiveness} Second-hand GPUs offer great price competitiveness.
Today's second-hand market prices for V100 16GB cards are now typically $\$ 60$ if one purchases in batch,
making them more than $100 \times$ cheaper than B200 GPUs (an 8x B200 system is around $\$ 600$K).
We acquired second-hand V100s and also sourced second-hand CPUs and motherboards to construct a compute \sysname{} using these devices. The cost of assembling this 128-V100 pod, composed entirely of used hardware, including all peripherals, is $\$32$K (as per March 2026 it further depreciated to $\$22$K).
Starting from this hardware configuration, we designed a highly optimized LLM serving engine. Our approach demonstrates that, despite individual V100 GPUs being significantly less powerful than H100/B200 GPUs according to datasheet specifications, careful software engineering can enable scaled-out computation with used GPUs to deliver LLM inference performance.

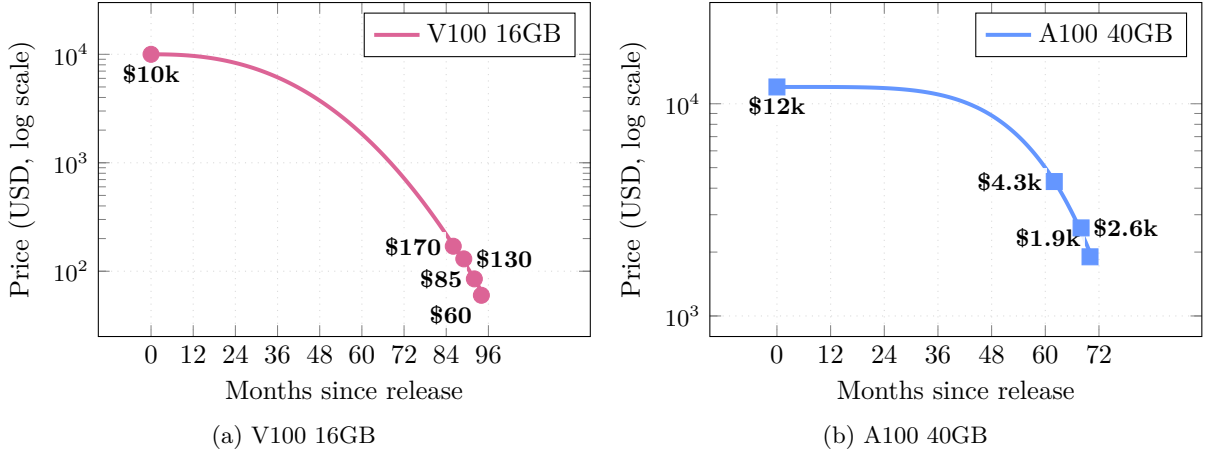
\begin{figure}[t]
    \centering
    \begin{subfigure}{.49\textwidth}
        \centering
        \begin{tikzpicture}
            \begin{axis}[
                width=\linewidth,
                height=6cm,
                xlabel={Months since release},
                ylabel={Price (USD, log scale)},
                xmin=-15, xmax=125,
                ymode=log,
                ymin=25, ymax=30000,
                xtick={0,12,24,36,48,60,72,84,96},
                ytick={100,1000,10000},
                yticklabel style={font=\small},
                grid=major,
                grid style={dotted, gray!30},
                legend pos=north east,
                mark options={solid},
                clip=false,
            ]

            \addplot[
                color={rgb,255:red,220;green,100;blue,150},
                line width=1.5pt,
                domain=0:94,
                samples=200,
                smooth,
            ] {10000.00 * exp(-pow(x/48.33, 2.4280))};

            \addplot[
                only marks,
                color={rgb,255:red,220;green,100;blue,150},
                mark=*,
                mark size=3pt,
            ] coordinates {
                (0, 10000)
                (86, 170)
                (89, 130)
                (92, 85)
                (94, 60)
            };
            \addlegendentry{V100 16GB}

            \node[below, font=\footnotesize\bfseries, fill=white, fill opacity=0.9, text opacity=1, inner sep=1.5pt, outer sep=2pt] at (axis cs:0,10000) {\$10k};
            \node[anchor=east, font=\footnotesize\bfseries, fill=white, fill opacity=0.9, text opacity=1, inner sep=1.5pt, outer sep=3pt] at (axis cs:86,170) {\$170};
            \node[anchor=west, font=\footnotesize\bfseries, fill=white, fill opacity=0.9, text opacity=1, inner sep=1.5pt, outer sep=3pt] at (axis cs:89,130) {\$130};
            \node[anchor=east, font=\footnotesize\bfseries, fill=white, fill opacity=0.9, text opacity=1, inner sep=1.5pt, outer sep=3pt] at (axis cs:92,85) {\$85};
            \node[anchor=north east, font=\footnotesize\bfseries, fill=white, fill opacity=0.9, text opacity=1, inner sep=1.5pt, outer sep=2pt] at (axis cs:94,60) {\$60};

            \end{axis}
        \end{tikzpicture}
        \caption{V100 16GB}
        \label{fig:gpu_price:v100}
    \end{subfigure}%
    \begin{subfigure}{.49\textwidth}
        \centering
        \begin{tikzpicture}
            \begin{axis}[
                width=\linewidth,
                height=6cm,
                xlabel={Months since release},
                ylabel={Price (USD, log scale)},
                xmin=-15, xmax=95,
                ymode=log,
                ymin=800, ymax=30000,
                xtick={0,12,24,36,48,60,72},
                ytick={1000,10000},
                yticklabel style={font=\small},
                grid=major,
                grid style={dotted, gray!30},
                legend pos=north east,
                mark options={solid},
                clip=false,
            ]

            \addplot[
                color={rgb,255:red,100;green,150;blue,255},
                line width=1.5pt,
                domain=0:70,
                samples=200,
                smooth,
            ] {11999.29 * exp(-pow(x/61.72, 4.6617))};

            \addplot[
                only marks,
                color={rgb,255:red,100;green,150;blue,255},
                mark=square*,
                mark size=3pt,
            ] coordinates {
                (0, 12000)
                (62, 4300)
                (68, 2600)
                (70, 1900)
            };
            \addlegendentry{A100 40GB}

            \node[below, font=\footnotesize\bfseries, fill=white, fill opacity=0.9, text opacity=1, inner sep=1.5pt, outer sep=2pt] at (axis cs:0,12000) {\$12k};
            \node[anchor=east, font=\footnotesize\bfseries, fill=white, fill opacity=0.9, text opacity=1, inner sep=1.5pt, outer sep=3pt] at (axis cs:62,4300) {\$4.3k};
            \node[anchor=west, font=\footnotesize\bfseries, fill=white, fill opacity=0.9, text opacity=1, inner sep=1.5pt, outer sep=3pt] at (axis cs:68,2600) {\$2.6k};
            \node[anchor=south east, font=\footnotesize\bfseries, fill=white, fill opacity=0.9, text opacity=1, inner sep=1.5pt, outer sep=2pt] at (axis cs:70,1900) {\$1.9k};

            \end{axis}
        \end{tikzpicture}
        \caption{A100 40GB}
        \label{fig:gpu_price:a100}
    \end{subfigure}
    \caption{Second-hand GPU price depreciation over time relative to their release dates. V100 16GB was released in May 2018 with a launch price of \$10,000, while A100 40GB was released in May 2020 at \$12,000. Data points show actual market prices from eBay listings. Solid curves show fitted stretched exponential decay models. V100 prices drop to \$60 (99.4\% depreciation) and A100 to \$1,900 (84.2\% depreciation) by March 2026. Log scale on the y-axis reveals the accelerating depreciation pattern.}
    \label{fig:gpu_price}
\end{figure}

\begin{figure}[!htbp]
    \centering
    \definecolor{barcolor1}{RGB}{100, 150, 255}  
    \definecolor{barcolor2}{RGB}{255, 140, 0}    
    \definecolor{barcolor3}{RGB}{100, 200, 100}  
    \definecolor{barcolor4}{RGB}{220, 100, 150}  
    \definecolor{barcolor5}{RGB}{150, 100, 220}  
    \definecolor{barcolor6}{RGB}{200, 200, 200}  

\begin{subfigure}{.5\textwidth}
    \centering
    \begin{tikzpicture}
      \begin{axis}[
        ymode = log,
        log origin = infty,
        ybar,
        bar width=6pt,
        width=6.5cm,
        height=6cm,
        ylabel={Tokens per Second (TPS)},
        symbolic x coords={0,10,20,30, 40},
        xtick={0,10,20},
        xticklabels={ShareGPT,Prefill-heavy,Decode-heavy},
        x tick label style={rotate=10},
        enlarge x limits=0.25,
        legend style={at={(0.5,-0.22)},anchor=north,legend columns=3},
        legend image code/.code={
          \draw[#1] (0cm,-0.2cm) rectangle (0.6cm,0.1cm);
        },
      ]
        \addplot+[barcolor1,fill=barcolor1!30,bar shift=-10pt,forget plot] coordinates {
           (0,9312)
        };
        \addplot+[barcolor1,pattern=north east lines,pattern color=barcolor1,draw=none,bar shift=-10pt,forget plot] coordinates {
          (0,9312)
        };
        \addplot+[barcolor4,fill=barcolor4!30,bar shift=-2.5pt,forget plot] coordinates {
           (0,4551)
        };
        \addplot+[barcolor4,pattern=north east lines,pattern color=barcolor4,draw=none,bar shift=-2.5pt,forget plot] coordinates {
          (0,4551)
        };
        \addplot+[barcolor3,fill=barcolor3!30,bar shift=5pt,forget plot] coordinates {
           (0,672)
        };
        \addplot+[barcolor3,pattern=north east lines,pattern color=barcolor3,draw=none,bar shift=5pt,forget plot] coordinates {
          (0,672)
        };
        \addplot+[barcolor5,fill=barcolor5!30,bar shift=12pt,forget plot] coordinates {
           (0,21504)
        };
        \addplot+[barcolor5,pattern=north east lines,pattern color=barcolor5,draw=none,bar shift=12pt,forget plot] coordinates {
          (0,21504)
        };
        \addplot+[barcolor2,fill=barcolor2!30,bar shift=19pt,forget plot] coordinates {
           (0,43911)
        };
        \addplot+[barcolor2,pattern=north east lines,pattern color=barcolor2,draw=none,bar shift=19pt,forget plot] coordinates {
          (0,43911)
        };

        \addplot+[barcolor1,fill=barcolor1!30,bar shift=-10pt,forget plot] coordinates {
           (10,29343)
        };
        \addplot+[barcolor1,pattern=crosshatch,pattern color=barcolor1,draw=none,bar shift=-10pt,forget plot] coordinates {
          (10,29343)
        };
        \addplot+[barcolor4,fill=barcolor4!30,bar shift=-2.5pt,forget plot] coordinates {
           (10,12451)
        };
        \addplot+[barcolor4,pattern=crosshatch,pattern color=barcolor4,draw=none,bar shift=-2.5pt,forget plot] coordinates {
          (10,12451)
        };
        \addplot+[barcolor3,fill=barcolor3!30,bar shift=5pt,forget plot] coordinates {
           (10,6998)
        };
        \addplot+[barcolor3,pattern=crosshatch,pattern color=barcolor3,draw=none,bar shift=5pt,forget plot] coordinates {
          (10,6998)
        };
        \addplot+[barcolor5,fill=barcolor5!30,bar shift=12pt,forget plot] coordinates {
           (10,223936)
        };
        \addplot+[barcolor5,pattern=crosshatch,pattern color=barcolor5,draw=none,bar shift=12pt,forget plot] coordinates {
          (10,223936)
        };
        \addplot+[barcolor2,fill=barcolor2!30,bar shift=19pt,forget plot] coordinates {
           (10,62651)
        };
        \addplot+[barcolor2,pattern=crosshatch,pattern color=barcolor2,draw=none,bar shift=19pt,forget plot] coordinates {
          (10,62651)
        };

        \addplot+[barcolor1,fill=barcolor1!30,bar shift=-10pt,forget plot] coordinates {
           (20,11509)
        };
        \addplot+[barcolor1,pattern=grid,pattern color=barcolor1,draw=none,bar shift=-10pt,forget plot] coordinates {
          (20,11509)
        };
        \addplot+[barcolor4,fill=barcolor4!30,bar shift=-2.5pt,forget plot] coordinates {
           (20,5919)
        };
        \addplot+[barcolor4,pattern=grid,pattern color=barcolor4,draw=none,bar shift=-2.5pt,forget plot] coordinates {
          (20,5919)
        };
        \addplot+[barcolor3,fill=barcolor3!30,bar shift=5pt,forget plot] coordinates {
           (20,254)
        };
        \addplot+[barcolor3,pattern=grid,pattern color=barcolor3,draw=none,bar shift=5pt,forget plot] coordinates {
          (20,254)
        };
        \addplot+[barcolor5,fill=barcolor5!30,bar shift=12pt,forget plot] coordinates {
           (20,8128)
        };
        \addplot+[barcolor5,pattern=grid,pattern color=barcolor5,draw=none,bar shift=12pt,forget plot] coordinates {
          (20,8128)
        };
        \addplot+[barcolor2,fill=barcolor2!30,bar shift=19pt,forget plot] coordinates {
           (20,26037)
        };
        \addplot+[barcolor2,pattern=grid,pattern color=barcolor2,draw=none,bar shift=19pt,forget plot] coordinates {
          (20,26037)
        };

        \addlegendimage{fill=barcolor1!30}
        \addlegendentry{1 H100}
        \addlegendimage{fill=barcolor4!30}
        \addlegendentry{1 A100}
        \addlegendimage{fill=barcolor3!30}
        \addlegendentry{4 V100}
        \addlegendimage{fill=barcolor5!30}
        \addlegendentry{128 V100}
        \addlegendimage{fill=barcolor2!30}
        \addlegendentry{1 B200}
        \addlegendimage{fill=black!30,pattern=north east lines,pattern color=black!60}
        \addlegendentry{ShareGPT}
        \addlegendimage{fill=black!30,pattern=crosshatch,pattern color=black!60}
        \addlegendentry{Prefill-heavy}
        \addlegendimage{fill=black!30,pattern=grid,pattern color=black!60}
        \addlegendentry{Decode-heavy}
      \end{axis}
    \end{tikzpicture}
    \caption{LLaMA 3.1-8B}
    \label{fig:comp:8b}
    \end{subfigure}%
\begin{subfigure}{.5\textwidth}
    \centering
    \begin{tikzpicture}
      \begin{axis}[
        ymode = log,
        log origin = infty,
        ybar,
        bar width=6pt,
        width=6.5cm,
        height=6cm,
        ylabel={~},
        symbolic x coords={0,10,20,30, 40},
        xtick={0,10,20},
        xticklabels={ShareGPT,Prefill-heavy,Decode-heavy},
        x tick label style={rotate=10},
        enlarge x limits=0.25,
        legend style={at={(0.5,-0.22)},anchor=north,legend columns=3},
        legend image code/.code={
          \draw[#1] (0cm,-0.2cm) rectangle (0.6cm,0.1cm);
        },
      ]
        \addplot+[barcolor1,fill=barcolor1!30,bar shift=-10pt,forget plot] coordinates {
           (0,5265)
        };
        \addplot+[barcolor1,pattern=north east lines,pattern color=barcolor1,draw=none,bar shift=-10pt,forget plot] coordinates {
          (0,5265)
        };
        \addplot+[barcolor4,fill=barcolor4!30,bar shift=-2.5pt,forget plot] coordinates {
           (0,3402)
        };
        \addplot+[barcolor4,pattern=north east lines,pattern color=barcolor4,draw=none,bar shift=-2.5pt,forget plot] coordinates {
          (0,3402)
        };
        \addplot+[barcolor3,fill=barcolor3!30,bar shift=5pt,forget plot] coordinates {
           (0,440)
        };
        \addplot+[barcolor3,pattern=north east lines,pattern color=barcolor3,draw=none,bar shift=5pt,forget plot] coordinates {
          (0,440)
        };
        \addplot+[barcolor5,fill=barcolor5!30,bar shift=12pt,forget plot] coordinates {
           (0,3520)
        };
        \addplot+[barcolor5,pattern=north east lines,pattern color=barcolor5,draw=none,bar shift=12pt,forget plot] coordinates {
          (0,3520)
        };
        \addplot+[barcolor2,fill=barcolor2!30,bar shift=19pt,forget plot] coordinates {
           (0,41857)
        };
        \addplot+[barcolor2,pattern=north east lines,pattern color=barcolor2,draw=none,bar shift=19pt,forget plot] coordinates {
          (0,41857)
        };

        \addplot+[barcolor1,fill=barcolor1!30,bar shift=-10pt,forget plot] coordinates {
           (10,19885)
        };
        \addplot+[barcolor1,pattern=crosshatch,pattern color=barcolor1,draw=none,bar shift=-10pt,forget plot] coordinates {
          (10,19885)
        };
        \addplot+[barcolor4,fill=barcolor4!30,bar shift=-2.5pt,forget plot] coordinates {
           (10,8642)
        };
        \addplot+[barcolor4,pattern=crosshatch,pattern color=barcolor4,draw=none,bar shift=-2.5pt,forget plot] coordinates {
          (10,8642)
        };
        \addplot+[barcolor3,fill=barcolor3!30,bar shift=5pt,forget plot] coordinates {
           (10,189)
        };
        \addplot+[barcolor3,pattern=crosshatch,pattern color=barcolor3,draw=none,bar shift=5pt,forget plot] coordinates {
          (10,189)
        };
        \addplot+[barcolor5,fill=barcolor5!30,bar shift=12pt,forget plot] coordinates {
           (10,1512)
        };
        \addplot+[barcolor5,pattern=crosshatch,pattern color=barcolor5,draw=none,bar shift=12pt,forget plot] coordinates {
          (10,1512)
        };
        \addplot+[barcolor2,fill=barcolor2!30,bar shift=19pt,forget plot] coordinates {
           (10,37045)
        };
        \addplot+[barcolor2,pattern=crosshatch,pattern color=barcolor2,draw=none,bar shift=19pt,forget plot] coordinates {
          (10,37045)
        };

        \addplot+[barcolor1,fill=barcolor1!30,bar shift=-10pt,forget plot] coordinates {
           (20,10372)
        };
        \addplot+[barcolor1,pattern=grid,pattern color=barcolor1,draw=none,bar shift=-10pt,forget plot] coordinates {
          (20,10372)
        };
        \addplot+[barcolor4,fill=barcolor4!30,bar shift=-2.5pt,forget plot] coordinates {
           (20,5366)
        };
        \addplot+[barcolor4,pattern=grid,pattern color=barcolor4,draw=none,bar shift=-2.5pt,forget plot] coordinates {
          (20,5366)
        };
        \addplot+[barcolor3,fill=barcolor3!30,bar shift=5pt,forget plot] coordinates {
           (20,317)
        };
        \addplot+[barcolor3,pattern=grid,pattern color=barcolor3,draw=none,bar shift=5pt,forget plot] coordinates {
          (20,317)
        };
        \addplot+[barcolor5,fill=barcolor5!30,bar shift=12pt,forget plot] coordinates {
           (20,2536)
        };
        \addplot+[barcolor5,pattern=grid,pattern color=barcolor5,draw=none,bar shift=12pt,forget plot] coordinates {
          (20,2536)
        };
        \addplot+[barcolor2,fill=barcolor2!30,bar shift=19pt,forget plot] coordinates {
           (20,24695)
        };
        \addplot+[barcolor2,pattern=grid,pattern color=barcolor2,draw=none,bar shift=19pt,forget plot] coordinates {
          (20,24695)
        };

        \addlegendimage{fill=barcolor1!30}
        \addlegendentry{8x H100}
        \addlegendimage{fill=barcolor4!30}
        \addlegendentry{8x A100}
        \addlegendimage{fill=barcolor3!30}
        \addlegendentry{16 V100}
        \addlegendimage{fill=barcolor5!30}
        \addlegendentry{128 V100}
        \addlegendimage{fill=barcolor2!30}
        \addlegendentry{8x B200}
        \addlegendimage{fill=black!30,pattern=north east lines,pattern color=black!60}
        \addlegendentry{ShareGPT}
        \addlegendimage{fill=black!30,pattern=crosshatch,pattern color=black!60}
        \addlegendentry{Prefill-heavy}
        \addlegendimage{fill=black!30,pattern=grid,pattern color=black!60}
        \addlegendentry{Decode-heavy}
      \end{axis}
    \end{tikzpicture}
    \caption{LLaMA 3.1-70B}
    \label{fig:comp:70b}
    \end{subfigure}

    \vspace{0.75cm}
    \centering
    \begin{tabular}{rllccp{5.5cm}}
        \toprule
        & \textbf{Configuration} & \textbf{Price} & \multicolumn{2}{c}{\textbf{Peak TPS/\$}} & \textbf{Notes} \\
        \cmidrule(lr){4-5}
        & & & \textbf{8B} & \textbf{70B} & \\
        \midrule
        \tikz\draw[barcolor1,fill=barcolor1!30] (0,0) rectangle (0.4cm,0.2cm); & 1 H100 & \$18.5K & 1.59 & - & Secondary market, GPU only \\
        \tikz\draw[barcolor1,fill=barcolor1!30] (0,0) rectangle (0.4cm,0.2cm); & 8x H100 (pod) & \$148K & - & 0.13 & Secondary market, GPUs only \\
        \tikz\draw[barcolor4,fill=barcolor4!30] (0,0) rectangle (0.4cm,0.2cm); & 1 A100 & \$5.8K & 2.15 & - & Secondary market, GPU only \\
        \tikz\draw[barcolor4,fill=barcolor4!30] (0,0) rectangle (0.4cm,0.2cm); & 8x A100 (pod) & \$46.4K & - & 0.19 & Secondary market, GPUs only \\
        \tikz\draw[barcolor2,fill=barcolor2!30] (0,0) rectangle (0.4cm,0.2cm); & 1 B200 & \$75K & 0.84 & - & Derived from full system MSRP \\
        \tikz\draw[barcolor2,fill=barcolor2!30] (0,0) rectangle (0.4cm,0.2cm); & 8x B200 (pod) & \$600K & - & 0.06 & Full system MSRP  \\
        \tikz\draw[barcolor3,fill=barcolor3!30] (0,0) rectangle (0.4cm,0.2cm); & 4 V100 (pod) & \$240 & 29.16 & - & Secondary market, GPUs only \\
        \tikz\draw[barcolor3,fill=barcolor3!30] (0,0) rectangle (0.4cm,0.2cm); & 16 V100 (pod) & \$960 & - & 0.20 & Secondary market, GPUs only \\
        \tikz\draw[barcolor5,fill=barcolor5!30] (0,0) rectangle (0.4cm,0.2cm); & 128 V100 (pod) & \$7.68K & 29.16 & 0.20 & GPUs only (Full system with peripherals is $\approx$\$22K) \\
        \bottomrule
    \end{tabular}

    \caption{Tokens per Second (TPS) for B200, H100, A100, and V100 devices evaluated across various workloads. Hardware configurations range from single instances to full 128-device \sysname{} pods, with associated capital costs and calculated Peak TPS per dollar included for economic comparison. The monetary cost used in calculation here contains only capital expenditure and does not take operational expenditure into account.}
\end{figure}

To quantify the price depreciation patterns, we collected second-hand market prices for NVIDIA V100 16GB and A100 40GB GPUs over time, relative to their release dates (May 2018 and May 2020, respectively).
We fitted these depreciation curves using a stretched exponential decay model: $P(t) = P_0 \cdot e^{-(t/\tau)^\beta}$, where $P_0$ is the initial launch price, $\tau$ represents the characteristic time scale, and $\beta$ is the stretching exponent that captures the rate of depreciation acceleration.
For V100, we obtained $\tau=48.3$ months and $\beta=2.43$ (R$^2$=1.000); for A100, $\tau=61.7$ months and $\beta=4.66$ (R$^2$=0.9997).
The values of $\beta > 1$ indicate accelerating depreciation, meaning GPUs lose value at an increasing rate as they age; this is a pattern consistent with technology obsolescence in the face of rapid hardware innovation cycles.
\Cref{fig:gpu_price} shows the fitted curves on a logarithmic price scale, revealing the non-linear decay pattern.
It is worth noting that batch purchase pricing data is unavailable when these devices are actively deployed in hyperscaler datacenters. Consequently, the x-axis in \Cref{fig:gpu_price} shows data points only after 3-4 years from release, when GPUs are phased out from large-scale production deployments.

\noindent
\textbf{Amortized/annualized embodied carbon} Re-utilizing second-hand GPUs offers a great reduction in the embodied carbon ($C_{\text{em}}$). This is because the manufacture of GPUs and host servers is an energy-intensive process that can result in a high volume of embodied carbon~\cite{li2023toward, faiz2023llmcarbon, li2024towards, tomlinson2024carbon, hewage2025aging}, as shown in \Cref{tab:cpa-table}. By re-deploying second-hand GPUs as a \sysname{}, this effectively offers a reduction in the annualized embodied carbon $C_{\text{em}}^{\text{annual}}(T)$ (\Cref{eq:annual}) by extending their life span.
Furthermore, by reusing GPUs, our approach can reduce the demand for manufacturing new GPUs of equivalent capacity, further reducing $C_{\text{em}}$ by means of circular economy.
As a secondary benefit, this reduced demand could also help alleviate the current GPU shortage. Deeper analysis of the modern accelerator supply chain is beyond the scope of this paper.

\noindent
\textbf{Operational carbon and costs} Second-hand GPUs may incur higher operational carbon ($C_{\text{opt}}$, \Cref{eq:copt}) emissions and costs if deployed without optimization.
Previous-generation GPUs are manufactured with older silicon technology nodes; for example, V100 uses TSMC 12nm while H100 uses TSMC N4. Later-generation GPUs are also equipped with higher-bandwidth HBM memory, V100 utilizes HBM2 whereas H100 deploys HBM3.
Advances in manufacturing nodes offer higher transistor density, improving energy efficiency, while larger and higher-bandwidth HBM also contributes to reduced operational costs. These energy efficiency improvements reduce both operational carbon and costs for newer-generation GPUs.
Furthermore, operational monetary cost is primarily determined by energy supply pricing, while operational carbon depends on the energy mix.
Lower carbon footprint requires energy mixes dominated by non-fossil fuel sources.
We made an interesting observation in this paper: cost-effectiveness and environmental sustainability both depend on grid pricing and energy supply mix, as inexpensive energy from low carbon sources is a critical component for making inference serving on second-hand devices both economically and environmentally viable.

\noindent
\textbf{Reliability of used hardware} Second-hand GPUs can exhibit increased fault rates. Since used GPUs have typically been deployed for a full product cycle, approximately 3 years from launch to decommission~\footnote{Though newer-generation GPUs are launched on roughly 2-year cycles, the typical deployment period from shipment to decommissioning is around 3 years, aligned with warranty periods.}, hardware reliability may be degraded. Previous analyses of GPU reliability~\cite{ostrouchov2020gpu, kokolis2024revisiting} indicate that most reliability issues originate from the GPU devices themselves, and system reliability can therefore decrease with extended GPU lifecycles. However, this can be addressed proactively through hardware binning: stress-testing used GPUs before deployment and removing units unsuitable for the target workload.
In our cluster construction, we conducted this binning process to ensure reliability.
Additionally, we implement software-level mitigation strategies. We employ redundant capacity with load balancing techniques to improve overall system reliability, and in practice we provision $10\%$ additional GPU devices to ensure cluster reliability.

\subsection{Building a GPU \sysname{}}

\begin{figure}[!t]
    \centering
\includegraphics[width=0.9\linewidth]{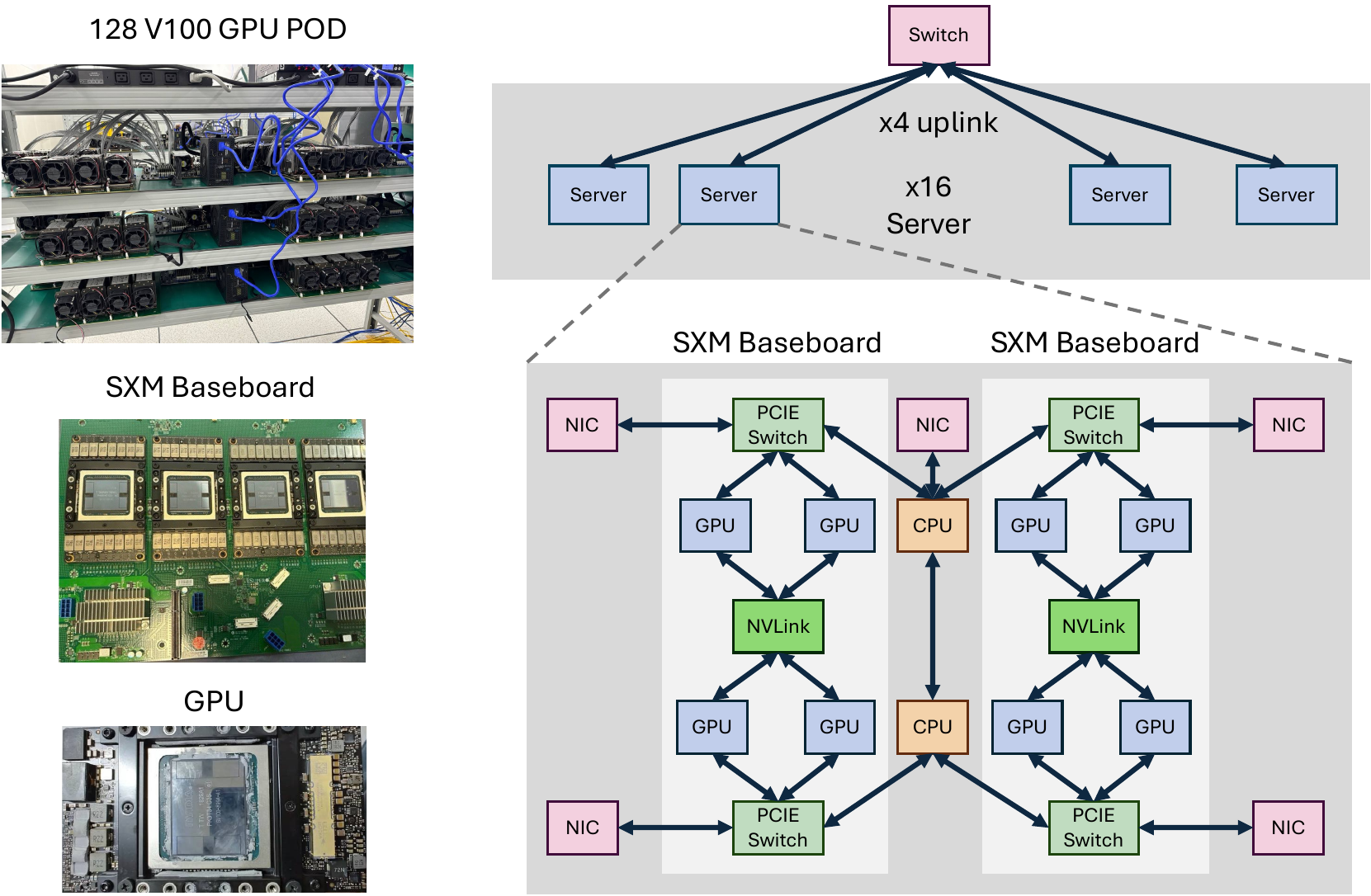}
    \caption{An illustration and real pictures of our used GPU pod configuration. The server topology and pod topology are shown with key components illustrated on the top. Each server node contains 8 V100 GPUs, and the GPU server pod contains 16 server nodes, providing a total of 128 V100 GPUs. Our cluster has two of these pods.}
    \label{fig:cluster-config}
\end{figure}

To demonstrate the potential of second-hand GPUs in serving LLMs, we assembled a GPU \sysname{} with used V100 GPUs as illustrated in \Cref{fig:cluster-config}, where images are taken from our actual hardware fleet. The pod consisted of $128$ V100 GPU devices, acquired at approximately $\$ 200$ per unit \footnote{The acquisition of these GPU devices was in May 2024 at $\$ 215$ and July 2024 at $\$ 170$, the current price of these V100 is at around $\$ 50$ to $\$ 60$.}.

The GPUs are mounted on specialized SXM baseboards that facilitate $300GB/s$ bidirectional NVLink connectivity between sets of four GPUs.
The SXM baseboard is equipped with two PCIe 3.0 switches, each linking two V100 GPUs with a single $100G$ Connect-X 5 network interface card (NIC).
Each server node consists of two such baseboards, which are individually connected to a Numa Node on a Dual Socket server board with an
Intel Xeon 6138 processor and $256GB$ DRAM.
All 4 NICs are linked to a $100Gbps$ RDMA fabric, and there is an additional $25Gbps$ Connectx-4 NIC directly attached to CPU1 to handle management and service traffic.
The detailed topology and configuration are shown in \Cref{fig:cluster-config}.
In total, the construction of the GPU pod costs around $\$ 32,000$ in 2024, including all server and switch components, which is roughly the same price as a single H100 GPU device at its MSRP in 2024. If one would like to replicate such a GPU cluster now in March 2026, it would cost only around $\$ 22,000$.
Also, not only the GPUs but all other components, such as motherboards and network interface cards, were purchased second-hand. The entire server pod consists solely of refurbished devices.
Because hardware prices are influenced by market conditions and the depreciation of older GPU devices, we use the prices available to us at the time of writing -- specifically, those observed in March 2026 -- for all subsequent analysis and evaluation.

\subsection{Scaled-out parallelization for the \sysname{}}

\begin{figure}
    \centering
    \includegraphics[width=0.9\linewidth]{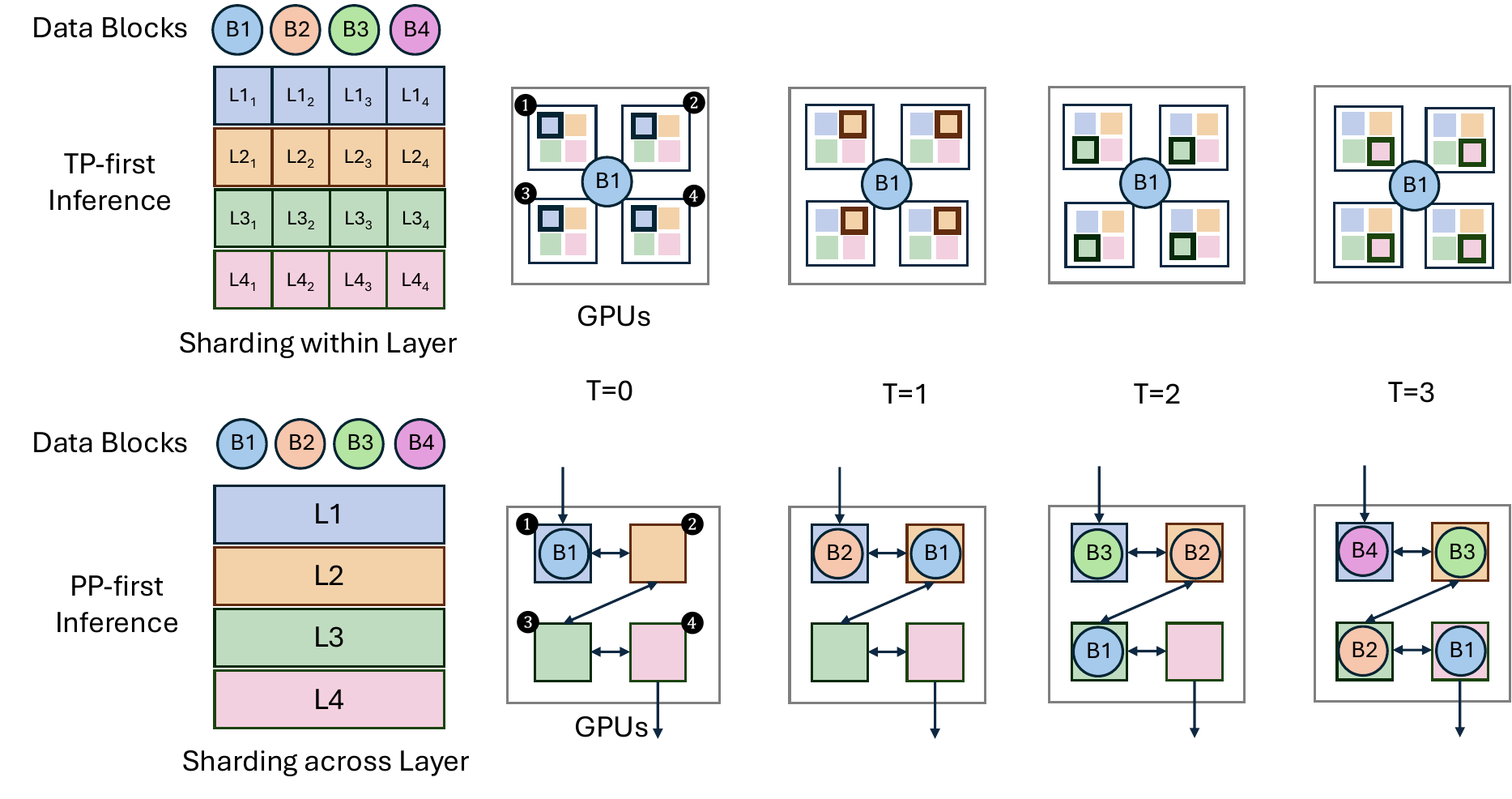}
    \caption{An illustration of the model parallelism strategy described with our approach. We use a simple example with inference on a 4 GPU node and a 4-layer model. For TP-first framework such as vLLM~\cite{kwon2023efficient}, they opted to utilize the high inter-device and compute-bandwidth ratio with a model sharding within layer. However, with our re-used devices, since our inter-device bandwidth and compute-bandwidth ratio are limited, for our approach (PP-first inference), we opt to utilize as much pipeline parallelism by layer as possible regardless of device topology. }
    \label{fig:pipeline_vs_tp}
\end{figure}

\begin{figure}[htbp]
    \centering
    \begin{tikzpicture}[scale=0.58, every node/.style={scale=0.70}]
        \definecolor{cpucolor}{RGB}{100, 150, 255}       
        \definecolor{computecolor}{RGB}{255, 140, 0}     
        \definecolor{gpucommcolor}{RGB}{220, 100, 150}   
        \definecolor{kernellaunch}{RGB}{200, 200, 200}   

        \node[anchor=west, font=\bfseries] at (-1.8, 4.2) {Naive Pipeline};

        \node[anchor=east] at (-0.2, 3.5) {GPU 0};
        \draw[thick, ->] (0, 3.5) -- (10, 3.5);
        \fill[computecolor!70] (0, 3.3) rectangle (1.0, 3.7);
        \node[white, font=\tiny\bfseries] at (0.5, 3.5) {W0};
        \fill[kernellaunch!70] (1.0, 3.3) rectangle (1.2, 3.7);
        \fill[gpucommcolor!70] (1.2, 3.3) rectangle (1.6, 3.7);
        \fill[computecolor!70] (2.5, 3.3) rectangle (3.5, 3.7);
        \node[white, font=\tiny\bfseries] at (3.0, 3.5) {W1};
        \fill[kernellaunch!70] (3.5, 3.3) rectangle (3.7, 3.7);
        \fill[gpucommcolor!70] (3.7, 3.3) rectangle (4.1, 3.7);
        \fill[computecolor!70] (5.0, 3.3) rectangle (6.0, 3.7);
        \node[white, font=\tiny\bfseries] at (5.5, 3.5) {W2};
        \fill[kernellaunch!70] (6.0, 3.3) rectangle (6.2, 3.7);
        \fill[gpucommcolor!70] (6.2, 3.3) rectangle (6.6, 3.7);

        \node[anchor=east] at (-0.2, 2.8) {GPU 1};
        \draw[thick, ->] (0, 2.8) -- (10, 2.8);
        \fill[gpucommcolor!70] (1.2, 2.6) rectangle (1.6, 3.0);
        \fill[computecolor!70] (1.6, 2.6) rectangle (2.6, 3.0);
        \node[white, font=\tiny\bfseries] at (2.1, 2.8) {W0};
        \fill[kernellaunch!70] (2.6, 2.6) rectangle (2.8, 3.0);
        \fill[gpucommcolor!70] (2.8, 2.6) rectangle (3.2, 3.0);
        \fill[gpucommcolor!70] (3.7, 2.6) rectangle (4.1, 3.0);
        \fill[computecolor!70] (4.1, 2.6) rectangle (5.1, 3.0);
        \node[white, font=\tiny\bfseries] at (4.6, 2.8) {W1};
        \fill[kernellaunch!70] (5.1, 2.6) rectangle (5.3, 3.0);
        \fill[gpucommcolor!70] (5.3, 2.6) rectangle (5.7, 3.0);
        \fill[gpucommcolor!70] (6.2, 2.6) rectangle (6.6, 3.0);
        \fill[computecolor!70] (6.6, 2.6) rectangle (7.6, 3.0);
        \node[white, font=\tiny\bfseries] at (7.1, 2.8) {W2};
        \fill[kernellaunch!70] (7.6, 2.6) rectangle (7.8, 3.0);
        \fill[gpucommcolor!70] (7.8, 2.6) rectangle (8.2, 3.0);

        \node[anchor=east] at (-0.2, 2.1) {GPU 2};
        \draw[thick, ->] (0, 2.1) -- (10, 2.1);
        \fill[gpucommcolor!70] (2.8, 1.9) rectangle (3.2, 2.3);
        \fill[computecolor!70] (3.2, 1.9) rectangle (4.2, 2.3);
        \node[white, font=\tiny\bfseries] at (3.7, 2.1) {W0};
        \fill[kernellaunch!70] (4.2, 1.9) rectangle (4.4, 2.3);
        \fill[gpucommcolor!70] (4.4, 1.9) rectangle (4.8, 2.3);
        \fill[gpucommcolor!70] (5.3, 1.9) rectangle (5.7, 2.3);
        \fill[computecolor!70] (5.7, 1.9) rectangle (6.7, 2.3);
        \node[white, font=\tiny\bfseries] at (6.2, 2.1) {W1};
        \fill[kernellaunch!70] (6.7, 1.9) rectangle (6.9, 2.3);
        \fill[gpucommcolor!70] (6.9, 1.9) rectangle (7.3, 2.3);
        \fill[gpucommcolor!70] (7.8, 1.9) rectangle (8.2, 2.3);
        \fill[computecolor!70] (8.2, 1.9) rectangle (9.2, 2.3);
        \node[white, font=\tiny\bfseries] at (8.7, 2.1) {W2};

        \node[anchor=east] at (-0.2, 1.4) {GPU 3};
        \draw[thick, ->] (0, 1.4) -- (10, 1.4);
        \fill[gpucommcolor!70] (4.4, 1.2) rectangle (4.8, 1.6);
        \fill[computecolor!70] (4.8, 1.2) rectangle (5.8, 1.6);
        \node[white, font=\tiny\bfseries] at (5.3, 1.4) {W0};
        \fill[kernellaunch!70] (5.8, 1.2) rectangle (6.0, 1.6);
        \fill[gpucommcolor!70] (6.0, 1.2) rectangle (6.4, 1.6);
        \fill[gpucommcolor!70] (6.9, 1.2) rectangle (7.3, 1.6);
        \fill[computecolor!70] (7.3, 1.2) rectangle (8.3, 1.6);
        \node[white, font=\tiny\bfseries] at (7.8, 1.4) {W1};
        \fill[kernellaunch!70] (8.3, 1.2) rectangle (8.5, 1.6);
        \fill[gpucommcolor!70] (8.5, 1.2) rectangle (8.9, 1.6);

        \draw[->, thick, gray!70] (0.5, 3.3) to[out=-45, in=135] (2.1, 3.0);
        \draw[->, thick, gray!70] (2.1, 2.6) to[out=-45, in=135] (3.7, 2.3);
        \draw[->, thick, gray!70] (3.7, 1.9) to[out=-45, in=135] (5.3, 1.6);

        \draw[->, thick, gray!70] (3.0, 3.3) to[out=-45, in=135] (4.6, 3.0);
        \draw[->, thick, gray!70] (4.6, 2.6) to[out=-45, in=135] (6.2, 2.3);
        \draw[->, thick, gray!70] (6.2, 1.9) to[out=-45, in=135] (7.8, 1.6);

        \draw[->, thick, gray!70] (5.5, 3.3) to[out=-45, in=135] (7.1, 3.0);
        \draw[->, thick, gray!70] (7.1, 2.6) to[out=-45, in=135] (8.7, 2.3);
        \draw[->, thick, gray!70] (8.7, 1.9) to[out=-45, in=135] (9.5, 1.6);

        \node[anchor=west, font=\bfseries] at (-1.8, 0.3) {Our Pipelining};

        \node[anchor=east] at (-0.2, -0.2) {GPU 0};
        \draw[thick, ->] (0, -0.2) -- (10, -0.2);
        \fill[computecolor!70] (0, -0.4) rectangle (1.0, 0.0);
        \node[white, font=\tiny\bfseries] at (0.5, -0.2) {W0};
        \fill[computecolor!70] (1.2, -0.4) rectangle (2.2, 0.0);
        \node[white, font=\tiny\bfseries] at (1.7, -0.2) {W1};
        \fill[computecolor!70] (2.4, -0.4) rectangle (3.4, 0.0);
        \node[white, font=\tiny\bfseries] at (2.9, -0.2) {W2};

        \node[anchor=east] at (-0.2, -0.65) {\small CPU};
        \draw[thick, ->] (0, -0.65) -- (10, -0.65);
        \fill[cpucolor!70] (0.9, -0.75) rectangle (1.3, -0.55);
        \fill[cpucolor!70] (2.1, -0.75) rectangle (2.5, -0.55);
        \fill[cpucolor!70] (3.3, -0.75) rectangle (3.7, -0.55);

        \node[anchor=east] at (-0.2, -1.15) {GPU 1};
        \draw[thick, ->] (0, -1.15) -- (10, -1.15);
        \fill[computecolor!70] (1.0, -1.35) rectangle (2.0, -0.95);
        \node[white, font=\tiny\bfseries] at (1.5, -1.15) {W0};
        \fill[computecolor!70] (2.2, -1.35) rectangle (3.2, -0.95);
        \node[white, font=\tiny\bfseries] at (2.7, -1.15) {W1};
        \fill[computecolor!70] (3.4, -1.35) rectangle (4.4, -0.95);
        \node[white, font=\tiny\bfseries] at (3.9, -1.15) {W2};

        \node[anchor=east] at (-0.2, -1.6) {\small CPU};
        \draw[thick, ->] (0, -1.6) -- (10, -1.6);
        \fill[cpucolor!70] (1.9, -1.7) rectangle (2.3, -1.5);
        \fill[cpucolor!70] (3.1, -1.7) rectangle (3.5, -1.5);
        \fill[cpucolor!70] (4.3, -1.7) rectangle (4.7, -1.5);

        \node[anchor=east] at (-0.2, -2.1) {GPU 2};
        \draw[thick, ->] (0, -2.1) -- (10, -2.1);
        \fill[computecolor!70] (2.0, -2.3) rectangle (3.0, -1.9);
        \node[white, font=\tiny\bfseries] at (2.5, -2.1) {W0};
        \fill[computecolor!70] (3.2, -2.3) rectangle (4.2, -1.9);
        \node[white, font=\tiny\bfseries] at (3.7, -2.1) {W1};
        \fill[computecolor!70] (4.4, -2.3) rectangle (5.4, -1.9);
        \node[white, font=\tiny\bfseries] at (4.9, -2.1) {W2};

        \node[anchor=east] at (-0.2, -2.55) {\small CPU};
        \draw[thick, ->] (0, -2.55) -- (10, -2.55);
        \fill[cpucolor!70] (2.9, -2.65) rectangle (3.3, -2.45);
        \fill[cpucolor!70] (4.1, -2.65) rectangle (4.5, -2.45);
        \fill[cpucolor!70] (5.3, -2.65) rectangle (5.7, -2.45);

        \node[anchor=east] at (-0.2, -3.05) {GPU 3};
        \draw[thick, ->] (0, -3.05) -- (10, -3.05);
        \fill[computecolor!70] (3.0, -3.25) rectangle (4.0, -2.85);
        \node[white, font=\tiny\bfseries] at (3.5, -3.05) {W0};
        \fill[computecolor!70] (4.2, -3.25) rectangle (5.2, -2.85);
        \node[white, font=\tiny\bfseries] at (4.7, -3.05) {W1};
        \fill[computecolor!70] (5.4, -3.25) rectangle (6.4, -2.85);
        \node[white, font=\tiny\bfseries] at (5.9, -3.05) {W2};

        \node[anchor=east] at (-0.2, -3.5) {\small CPU};
        \draw[thick, ->] (0, -3.5) -- (10, -3.5);
        \fill[cpucolor!70] (3.9, -3.6) rectangle (4.3, -3.4);
        \fill[cpucolor!70] (5.1, -3.6) rectangle (5.5, -3.4);
        \fill[cpucolor!70] (6.3, -3.6) rectangle (6.7, -3.4);

        \draw[->, thick, gray!70] (0.5, -0.4) to[out=-90, in=90] (1.5, -0.95);
        \draw[->, thick, gray!70] (1.5, -1.35) to[out=-90, in=90] (2.5, -1.9);
        \draw[->, thick, gray!70] (2.5, -2.3) to[out=-90, in=90] (3.5, -2.85);

        \draw[->, thick, gray!70] (1.7, -0.4) to[out=-90, in=90] (2.7, -0.95);
        \draw[->, thick, gray!70] (2.7, -1.35) to[out=-90, in=90] (3.7, -1.9);
        \draw[->, thick, gray!70] (3.7, -2.3) to[out=-90, in=90] (4.7, -2.85);

        \draw[->, thick, gray!70] (2.9, -0.4) to[out=-90, in=90] (3.9, -0.95);
        \draw[->, thick, gray!70] (3.9, -1.35) to[out=-90, in=90] (4.9, -1.9);
        \draw[->, thick, gray!70] (4.9, -2.3) to[out=-90, in=90] (5.9, -2.85);

        \node at (5, -4.2) {Time};

        \fill[computecolor!70] (10.3, 1.0) rectangle (10.7, 1.2);
        \node[anchor=west] at (10.8, 1.1) {GPU Compute};

        \fill[cpucolor!70] (10.3, 0.4) rectangle (10.7, 0.6);
        \node[anchor=west] at (10.8, 0.5) {CPU Send/Recv};

        \fill[gpucommcolor!70] (10.3, -0.2) rectangle (10.7, 0.0);
        \node[anchor=west] at (10.8, -0.1) {GPU Comm (NCCL)};

        \fill[kernellaunch!70] (10.3, -0.8) rectangle (10.7, -0.6);
        \node[anchor=west] at (10.8, -0.7) {Kernel Launch};

    \end{tikzpicture}
    \caption{An illustration of our pipelining compared to a naive pipelining strategy in a 4 GPU setting for inference. In the Naive Pipeline case, GPUs manage communication via NCCL, this requires a kernel launch and GPU sends/receives (blocking both GPUs). This creates a large idle time for the GPUs. In our pipelining approach, CPUs manage all communication asynchronously, overlapping with GPU compute, this requires only minimal buffering time in between GPU compute. Workload labels (W0-W2) and curved arrows show data flow across GPUs. GPUs remain dedicated to computation, achieving higher utilization and throughput.}
    \label{fig:pipeline}
\end{figure}
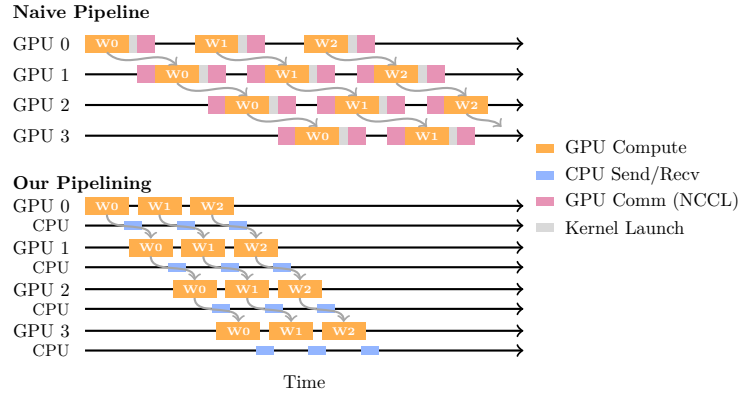

Older GPU devices commonly face the following constraints:
\begin{itemize}
    \item Limited HBM memory capacity;
    \item Limited HBM memory bandwidth;
    \item Reduced computing capability (lower FLOPs/second);
    \item Smaller inter-GPU communication bandwidth (slower NVLink).
\end{itemize}

The constraints of HBM memory and computing capability can be mitigated with scaled-out computing, which involves utilizing a greater number of GPU devices.
Scaled-out inference, in our design, is supported mainly through aggressive pipeline parallelism. Pipeline parallelism enables scaling by distributing different model segments across different GPUs, adhering to a predetermined parallelization strategy.

In addition to the common data parallelism, existing LLM serving software tools use mainly three forms of model parallelisms, namely tensor parallelism (TP), sequence parallelism (SP) and pipeline parallelism (PP) \footnote{There are also other forms of parallelism such as expert parallelism (for MoE models), but these are beyond the scope of this paper.}.
Frameworks like vLLM \cite{kwon2023efficient} and SGL \cite{zheng2024sglang} typically employ a mixed parallelization plan.
They use a heuristic that conducts intra-node level tensor parallelism first and further expands to pipeline or sequence parallelism. This effectively means that each layer of the model is divided using tensor parallelism within each device and requires a scatter and a gather communication operation, as shown in \Cref{fig:pipeline_vs_tp}, to communicate the inference inputs and results with intermediate partitions of a layer. This strategy demands a high GPU-GPU interconnect bandwidth between TP devices, which could be more prone to reliability issues and incurs increased communication costs with second-hand GPUs.
This mixed parallelization plan is efficient with cutting-edge GPUs like the H100 and B200 for two reasons: first, as detailed in \Cref{tab:cpa-table}, devices such as the H100 possess good local inter-GPU communication links; second, these devices are predominantly memory-bound, exhibiting a large compute-to-bandwidth ratio (such as 0.29 TFLOPs/GB for H100). This is related to semiconductor scaling; the growth in compute capabilities (FLOPs/s) has considerably outpaced that of memory interconnects (GB/s). Using tensor parallelism is a good strategy to harness more HBM bandwidth and achieve more efficient LLM serving.

This design choice, however, is less attractive for second-hand GPUs, which have a reduced inter-GPU bandwidth and, more critically, are not as memory-bound as these new GPUs.
To address this, we developed a \textbf{pipeline-first parallelization strategy} for second-hand GPUs, shown in \Cref{fig:pipeline}.
Our approach supports a novel device-level pipeline-first parallelism, where each device operates as an individual pipeline stage, diverging from traditional node-level pipelining \footnote{Commonly, 4-8 devices would form a compute node, as illustrated in \Cref{fig:cluster-config}, and pipelining happens at this node level.}.
This device-level pipelining is designed to accommodate at least one layer per GPU device, enabling a maximum pipeline parallelism and creating an exceptionally deep compute pipeline. This approach enabled us to scale up with a greater number of second-hand devices, circumventing the small VRAM and limited computational power of individual second-hand GPUs.
With this device-level pipelining, we can achieve up to 80 pipeline stages to fit an LLaMA-70B model on V100 devices, which is a significantly deeper pipeline compared to the one using vLLM \cite{kwon2023efficient}.

This particular optimization choice also accommodates the lower NVLink bandwidth in second-hand GPUs by only transferring the forward result after a layer (\eg{}~layer outputs), saving the communication cost for gather-scatter in TP. The amount of bandwidth required across GPUs is only 3.125 GB/s which is well below the cross-device bandwidth that the custom-built V100 GPU pod can offer.
In our design, we also support a mixed parallelization plan by optionally using tensor parallelism in the final phase, as these devices are inherently less memory bound (lower compute-to-bandwidth ratio shown in \Cref{tab:cpa-table}). In essence, frameworks such as vLLM \cite{kwon2023efficient} prioritize tensor parallelism, whereas our approach operates in a pipeline parallelism first manner when both are operating in mixed parallelization modes.
However, a deep device-level pipeline comes with the cost of increased cross-device communication to forward the intermediate results.
In a naively implemented pipeline parallel strategy (Naive Pipeline), computed results must be transmitted to subsequent devices, introducing communication latency between each pipeline stage.
As shown in \Cref{fig:pipeline}, this results in communication delays between each pipeline stage.
The common strategy is to establish such communication through the GPUs, which requires the GPUs to stop doing compute and launch a communication kernel (the NCCL and Kernel Launch block shown in \Cref{fig:pipeline}), which introduces extra communication latency into the pipeline and would force GPUs in the pipeline to stay idle from doing useful computation.
It is worth noting that this is also a scenario of TP inference with node-level pipelines.
To address this, we propose and have implemented an asynchronous data transaction optimization on the host (CPU) side. As \Cref{fig:pipeline} outlines, this optimization enables pre-fetching of data inputs for the ($t+1$)th timestep while the GPU is computing the $t$-th timestep results. Consequently, this prefetch masking hides the data transaction time and frees up the GPUs to do more useful work and eliminate their idle computational intervals.

\begin{table}[!t]
\caption{Hardware characterization and embodied carbon for GPUs. The die size and VRAM size data are obtained from datasheets and the manufacture CPA values are from~\cite{faiz2023llmcarbon} and \cite{bhagavathula2024understanding}. The VRAM embodied carbon is obtained following the unit CO2e shown in~\cite{li2024towards}. Given that a single deployment cycle is typically 3 years, we consider the embodied carbon for three generations of devices over potential deployment durations of 3, 5, and 8 years. For the B200 GPU, since the data are not available, we deduced the data using closest possible process node for the embodied carbon. These deduced results are marked with an $\ast$.}
\label{tab:cpa-table}
\begin{tabular}{@{}cccccc@{}}
\toprule
\textbf{Categories} & \textbf{Units} & \textbf{B200} & \textbf{H100 80GB} & \textbf{A100 80GB} & \textbf{V100 16GB} \\ \midrule
\multicolumn{6}{c}{Hardware characteristics}     \\ \midrule
Compute Capability  & TFLOPs/s       & 2250 (BF16)   & 989 (BF16)    & 312 (BF16)        & 125 (FP16)        \\
HBM BW              & GB/s      & 8000    & 3350    & 1555     & 900       \\
Compute to BW ratio & TFLOPs/GB & 0.28    & 0.29    & 0.2      & 0.13      \\
Inter-GPU BW        & GB/s      & 1800      & 900     & 600      & 300       \\
 \midrule
\multicolumn{6}{c}{Logic die embodied carbon}              \\ \midrule
Manufacture Process & -       & TSMC 4NP & TSMC N4 & TSMC 7nm & TSMC 12nm \\
Silicon Area        & mm2       & 1600      & 815     & 826      & 815       \\
CO2e Per Area       & kg/cm2    & 1.8* & 1.8     & 1.6      & 1.2       \\
CO2e in Total       & kg        & 28.8* & 14.67   & 13.21    & 9.78      \\ \midrule
\multicolumn{6}{c}{VRAM embodied carbon}                 \\ \midrule
VRAM Type           & -       & HBM3e   & HBM3    & HBM2e    & HBM2      \\
Size           & GB        & 192      & 80      & 80       & 16        \\
Unit CO2e      & kg/GB     & 0.24* & 0.24    & 0.24     & 0.28      \\
Total CO2e          & kg        & 46.08* & 19.2    & 19.2     & 4.48      \\ \midrule
\multicolumn{6}{c}{Equivalent yearly embodied carbon}   \\ \midrule
Carbon 3-year         & kg/yr       & 24.96* & 11.21    & 10.77     & 4.73      \\
Carbon 5-year         & kg/yr       & 12.48* & 6.74    & 6.48    & 2.80      \\
Carbon 8-year         & kg/yr      & 8.32* & 4.20    & 3.98     & 1.73  \\
\bottomrule
\end{tabular}

\end{table}

\section{Results}\label{sec:results}
\subsection{LLM inference service quality}

To evaluate whether second-hand GPUs can deliver production-quality LLM inference, we benchmark our custom serving engine against state-of-the-art systems running on current-generation hardware.

We focus on a metric that determines the viability of inference deployments: Tokens Per Second (TPS). This metric directly impacts user experience and operational costs, as modern LLM API providers charge per token generated. For the 8B model, our 128-V100 cluster achieves around 224K TPS on the Prefill-heavy workload, demonstrating that scaled-out second-hand hardware can deliver competitive throughput despite individual device limitations. For the 70B model, while absolute throughput is lower due to the larger model size, the pipeline-first parallelization strategy enables our V100 cluster to serve these models effectively, which would otherwise be impossible on individual older GPUs.

We evaluate performance across three representative workload categories, as illustrated in \Cref{fig:comp:8b,fig:comp:70b}. First, ShareGPT \citep{patel2024splitwise} represents real-world conversational traces with mixed input/output lengths typical of chatbot deployments. Second, Prefill-heavy workloads simulate compute-intensive scenarios such as code generation, where models process longer input contexts and generate substantial output. Third, Decode-heavy workloads emphasize rapid token generation with shorter prompts, common in conversational AI applications. This diversity ensures our evaluation captures the full spectrum of inference scenarios encountered in production deployments.
The details about each of these different workloads are summarized in \Cref{appendix:workload}.
As shown in \Cref{tab:cost-8b:prefill} and \Cref{fig:comp:8b}, for the 8B parameter model, our 128-V100 cluster demonstrates compelling performance. On the Prefill-heavy workload, the cluster achieves 223K TPS, substantially outperforming even the latest B200 GPU (around 62K TPS) through aggressive pipeline parallelization.
On ShareGPT, the cluster delivers 21.5K TPS compared to B200's 44K TPS.
This demonstrates that despite the individual limitations of V100 GPUs -- lower memory bandwidth (900 GB/s vs 3,350 GB/s for H100) and reduced compute capability in dense FP16 (125 TFLOPs vs 989 TFLOPs) -- our pipeline-first parallelization strategy enables competitive aggregate performance through scale-out computing.

The 70B parameter model presents a more demanding inference workload, requiring substantially more memory and compute resources. As detailed in \Cref{tab:cost-70b:prefill} and \Cref{fig:comp:70b}, our 128-V100 cluster achieves 1.5K TPS on the Prefill-heavy workload, compared to B200 8-GPU pod's 37K TPS and H100 8-GPU pod's 20K TPS, and the capital expenditure of 128-V100 is $78 \times$ less. 
While the absolute throughput is lower, this result is significant: the 70B model cannot fit on a single V100 (16GB memory), making individual V100s entirely unsuitable for this workload without our device-level pipeline parallelism approach.
The fact that we achieve any throughput at all -- let alone an order of the magnitude performance reduction -- demonstrates the effectiveness of our parallelization strategy in enabling second-hand hardware to serve models that would otherwise be impossible to deploy on older devices.
Our V100 systems, whether with 16-GPU or 128-GPU configurations, exhibit comparable and superior cost efficiency to the new B200 8-GPU pod, \textbf{particularly in regions with lower energy prices}. Additionally, the V100's greater availability in the supply chain and its reduced initial capital expenditure make it a more accessible option for assembly.
It should be noted that the A100 80GB model, as shown in \Cref{fig:comp:70b}, emerges as the most cost-effective option over an N-year time period. The A100-80GB system is calculated from a second-hand construction setup -- this reinforces our argument in favor of the superior cost efficiency of what we term ``\textit{\sysname{}}''.

A key observation from \Cref{fig:comp:8b,fig:comp:70b} is that our approach exhibits better scaling efficiency with larger models. For the 8B model, we achieve approximately 7.6$\times$ throughput improvement over a 4-V100 pod when scaling to 128 GPUs.
For the 70B model, where the deeper pipeline is essential rather than optional, the benefits of our device-level parallelization become more pronounced. This scaling behaviour validates our design choice: pipeline-first parallelization is particularly advantageous for larger models and greater device counts, precisely the scenario where second-hand GPUs offer the most compelling economic advantages through scale-out deployment.

\subsection{Cost and carbon footprint analysis}

\Cref{tab:cost-8b:prefill} and \Cref{tab:cost-70b:prefill} present our main findings, demonstrating the quality of inference service, the monetary costs, and the carbon footprint associated with the use of second-hand V100 GPU devices. For all results in these tables, we follow similar statistics with the conversation dataset published by Azure LLM~\cite{patel2024splitwise}, and consider an input length of 100 tokens generating 1024 output tokens. This typical measurement setup serves as a representation of human-AI chatbot dialogues. 
For the baseline comparisons, B200 benchmarks are conducted using vLLM v0.17, while H100 and A100 results are obtained using vLLM v0.6\footnote{H100 and A100 results were profiled at the time of finishing constructing the cluster, during the 1-year real-serving, later versions of vLLM and Blackwell devices became available.}{}; V100 results are from our custom serving system.

We present inference performance metrics, including Decode Throughput in Tokens Per Second (TPS), for both 8B and 70B LLaMA model \cite{dubey2024llama} inference. Tokens per Second (TPS) indicates system performance. Current LLM API providers charge based on each input and output token, so a higher TPS implies the potential for greater profits.
As illustrated in \Cref{tab:cost-8b:prefill}, a 128-V100 pod (\$7.68K in GPU cost) achieves approximately $3.6\times$ the throughput of a single B200 GPU (\$75K), while costing roughly $10\times$ less in hardware acquisition.
\Cref{tab:cost-70b:prefill}, which details the 70B model, shows that the 128-V100 pod achieves approximately $3.2\times$ the cost-effectiveness of a B200 system when considering only the capital expenditure required for hardware purchase, measured as throughput per dollar of GPU cost. It is important to note that our approach has a greater benefit as the model size increases, due to the pipeline-first parallelization strategy we've implemented.
This strategy is designed for scale-out computing and is naturally more advantageous with larger models and a higher number of devices.

We present the operational cost in \Cref{tab:cost-8b:prefill} and \Cref{tab:cost-70b:prefill}, which are closely linked to electricity prices. We consider four representative energy supplies from different geographical locations with varying mixes of renewable energy sources. Specifically, we consider:
\begin{itemize}
    \item US energy supply mix: $20\%$ renewable energy
    \item China energy supply mix: $30\%$ renewable energy
    \item Brazil energy supply mix: $90\%$ renewable energy
    \item Sole energy supply from wind farms in China: $100\%$ renewable energy
\end{itemize}
The fourth option ($100\%$ renewable) is an ideal setup, and even in this case, there is a residual operational carbon emission of approximately $10$\,g/kWh due to lifecycle emissions from renewable infrastructure.
Taking the official electricity unit costs from these regions (more details are available in \Cref{app:energy-grid}) we report operational expenses in terms of USD per million tokens (MT) generated.

It is evident in both \Cref{tab:cost-8b:prefill} and \Cref{tab:cost-70b:prefill}, state-of-the-art GPUs (like H100s) show significantly better operational cost mainly due to the silicon technology scaling. However, it is worth mentioning that existing LLM API companies, like Together AI, offer inferences for 8-billion-parameter models at 0.10 USD/MT and for 70-billion-parameter models at 0.88 USD/MT\footnote{These numbers are taken from inference cost of Llama3 8B Instruct Lite and Llama3.3 70B on \href{https://www.together.ai/pricing}{together.ai pricing}}. Even with heightened operational costs, systems based on V100 GPUs can still be extremely profitable.

\begin{figure*}[!t]
    \centering
    \definecolor{barcolor1}{RGB}{100, 150, 255}  
    \definecolor{barcolor2}{RGB}{255, 140, 0}    
    \definecolor{barcolor4}{RGB}{220, 100, 150}  
    \definecolor{barcolor5}{RGB}{150, 100, 220}  

    \pgfplotsset{
      costpanel/.style={
        log origin=infty,
        width=4.2cm,
        height=3.6cm,
        symbolic x coords={3-yr,5-yr,8-yr},
        xtick=data,
        enlarge x limits=0.15,
        x tick label style={font=\tiny},
        y tick label style={font=\tiny},
        yticklabel style={
          /pgf/number format/.cd,
          fixed,
          fixed zerofill,
          precision=2,
        },
        title style={font=\scriptsize, yshift=-2pt},
        ylabel style={font=\tiny},
        major grid style={line width=.1pt, draw=gray!20},
        ymajorgrids=true,
        scaled y ticks=false,
        legend style={font=\tiny},
      }
    }

    \begin{subfigure}{.5\textwidth}
        \centering
        \begin{tikzpicture}
          \begin{axis}[costpanel,
            title={China (30\%)},
            ylabel={USD/MT},
            ytick={0.001, 0.005, 0.01},
            ymin=0.0004, ymax=0.012]
            \addplot[barcolor2,thick,mark=square*,mark size=1.5pt]
              coordinates {(3-yr,0.01) (5-yr,0.01) (8-yr,0.0049)};
            \addplot[barcolor1,thick,mark=*,mark size=1.5pt]
              coordinates {(3-yr,0.01) (5-yr,0.0042) (8-yr,0.0027)};
            \addplot[barcolor4,thick,mark=triangle*,mark size=2pt]
              coordinates {(3-yr,0.01) (5-yr,0.0033) (8-yr,0.0022)};
            \addplot[barcolor5,thick,mark=diamond*,mark size=2pt]
              coordinates {(3-yr,0.0008) (5-yr,0.0007) (8-yr,0.0006)};
          \end{axis}
        \end{tikzpicture}%
        \begin{tikzpicture}
          \begin{axis}[costpanel,
            title={Wind (100\%)},
            ytick={0.001, 0.005, 0.01},
            yticklabels={},
            ymin=0.0004, ymax=0.012]
            \addplot[barcolor2,thick,mark=square*,mark size=1.5pt]
              coordinates {(3-yr,0.01) (5-yr,0.01) (8-yr,0.0049)};
            \addplot[barcolor1,thick,mark=*,mark size=1.5pt]
              coordinates {(3-yr,0.01) (5-yr,0.0043) (8-yr,0.0028)};
            \addplot[barcolor4,thick,mark=triangle*,mark size=2pt]
              coordinates {(3-yr,0.01) (5-yr,0.0035) (8-yr,0.0024)};
            \addplot[barcolor5,thick,mark=diamond*,mark size=2pt]
              coordinates {(3-yr,0.0010) (5-yr,0.0009) (8-yr,0.0008)};
          \end{axis}
        \end{tikzpicture}

        \vspace{0.08cm}

        \begin{tikzpicture}
          \begin{axis}[costpanel,
            title={US (20\%)},
            ylabel={USD/MT},
            ytick={0.001, 0.005, 0.01},
            ymin=0.0004, ymax=0.012]
            \addplot[barcolor2,thick,mark=square*,mark size=1.5pt]
              coordinates {(3-yr,0.01) (5-yr,0.01) (8-yr,0.01)};
            \addplot[barcolor1,thick,mark=*,mark size=1.5pt]
              coordinates {(3-yr,0.01) (5-yr,0.0046) (8-yr,0.0031)};
            \addplot[barcolor4,thick,mark=triangle*,mark size=2pt]
              coordinates {(3-yr,0.01) (5-yr,0.0040) (8-yr,0.0029)};
            \addplot[barcolor5,thick,mark=diamond*,mark size=2pt]
              coordinates {(3-yr,0.0017) (5-yr,0.0016) (8-yr,0.0015)};
          \end{axis}
        \end{tikzpicture}%
        \begin{tikzpicture}
          \begin{axis}[costpanel,
            title={Brazil (90\%)},
            ytick={0.001, 0.005, 0.01},
            yticklabels={},
            ymin=0.0004, ymax=0.012]
            \addplot[barcolor2,thick,mark=square*,mark size=1.5pt]
              coordinates {(3-yr,0.01) (5-yr,0.01) (8-yr,0.01)};
            \addplot[barcolor1,thick,mark=*,mark size=1.5pt]
              coordinates {(3-yr,0.01) (5-yr,0.0048) (8-yr,0.0033)};
            \addplot[barcolor4,thick,mark=triangle*,mark size=2pt]
              coordinates {(3-yr,0.01) (5-yr,0.0044) (8-yr,0.0033)};
            \addplot[barcolor5,thick,mark=diamond*,mark size=2pt]
              coordinates {(3-yr,0.0022) (5-yr,0.0021) (8-yr,0.0020)};
          \end{axis}
        \end{tikzpicture}

        \vspace{0.15cm}

        {\scriptsize
          \tikz[baseline=-0.5ex]{\draw[barcolor2,thick] (0,0.09cm) -- (0.35cm,0.09cm);
            \fill[barcolor2] (0.125cm,0.04cm) rectangle (0.225cm,0.14cm);}~1 B200
          \enspace
          \tikz[baseline=-0.5ex]{\draw[barcolor1,thick] (0,0.09cm) -- (0.35cm,0.09cm);
            \fill[barcolor1] (0.175cm,0.09cm) circle (0.05cm);}~1 H100
          \enspace
          \tikz[baseline=-0.5ex]{\draw[barcolor4,thick] (0,0.09cm) -- (0.35cm,0.09cm);
            \fill[barcolor4] (0.175cm,0.02cm) -- (0.24cm,0.09cm) -- (0.175cm,0.16cm) -- (0.11cm,0.09cm) -- cycle;}~1 A100
          \enspace
          \tikz[baseline=-0.5ex]{\draw[barcolor5,thick] (0,0.09cm) -- (0.35cm,0.09cm);
            \fill[barcolor5] (0.175cm,0.02cm) -- (0.24cm,0.09cm) -- (0.175cm,0.16cm) -- (0.11cm,0.09cm) -- cycle;}~128 V100
        }
        {\centering\caption{LLaMA 3.1-8B (Prefill-heavy)}\label{fig:cost-8b}\par}
    \end{subfigure}\hspace{-4pt}%
    \begin{subfigure}{.5\textwidth}
        \centering
        \vspace{0.05cm}
        \begin{tikzpicture}
          \begin{axis}[costpanel,
            title={China (30\%)},
            ylabel={USD/MT},
            ytick={0.03, 0.1, 0.3},
            ymin=0.02, ymax=0.4]
            \addplot[barcolor2,thick,mark=square*,mark size=1.5pt]
              coordinates {(3-yr,0.26) (5-yr,0.16) (8-yr,0.10)};
            \addplot[barcolor1,thick,mark=*,mark size=1.5pt]
              coordinates {(3-yr,0.16) (5-yr,0.09) (8-yr,0.06)};
            \addplot[barcolor4,thick,mark=triangle*,mark size=2pt]
              coordinates {(3-yr,0.10) (5-yr,0.06) (8-yr,0.04)};
            \addplot[barcolor5,thick,mark=diamond*,mark size=2pt]
              coordinates {(3-yr,0.07) (5-yr,0.06) (8-yr,0.05)};
          \end{axis}
        \end{tikzpicture}%
        \begin{tikzpicture}
          \begin{axis}[costpanel,
            title={Wind (100\%)},
            ytick={0.03, 0.1, 0.3},
            yticklabels={},
            ymin=0.02, ymax=0.4]
            \addplot[barcolor2,thick,mark=square*,mark size=1.5pt]
              coordinates {(3-yr,0.26) (5-yr,0.16) (8-yr,0.10)};
            \addplot[barcolor1,thick,mark=*,mark size=1.5pt]
              coordinates {(3-yr,0.16) (5-yr,0.10) (8-yr,0.06)};
            \addplot[barcolor4,thick,mark=triangle*,mark size=2pt]
              coordinates {(3-yr,0.10) (5-yr,0.06) (8-yr,0.04)};
            \addplot[barcolor5,thick,mark=diamond*,mark size=2pt]
              coordinates {(3-yr,0.09) (5-yr,0.08) (8-yr,0.07)};
          \end{axis}
        \end{tikzpicture}

        \vspace{0.08cm}

        \begin{tikzpicture}
          \begin{axis}[costpanel,
            title={US (20\%)},
            ylabel={USD/MT},
            ytick={0.03, 0.1, 0.3},
            ymin=0.02, ymax=0.4]
            \addplot[barcolor2,thick,mark=square*,mark size=1.5pt]
              coordinates {(3-yr,0.26) (5-yr,0.16) (8-yr,0.10)};
            \addplot[barcolor1,thick,mark=*,mark size=1.5pt]
              coordinates {(3-yr,0.16) (5-yr,0.10) (8-yr,0.07)};
            \addplot[barcolor4,thick,mark=triangle*,mark size=2pt]
              coordinates {(3-yr,0.11) (5-yr,0.07) (8-yr,0.05)};
            \addplot[barcolor5,thick,mark=diamond*,mark size=2pt]
              coordinates {(3-yr,0.15) (5-yr,0.14) (8-yr,0.13)};
          \end{axis}
        \end{tikzpicture}%
        \begin{tikzpicture}
          \begin{axis}[costpanel,
            title={Brazil (90\%)},
            ytick={0.03, 0.1, 0.3},
            yticklabels={},
            ymin=0.02, ymax=0.4]
            \addplot[barcolor2,thick,mark=square*,mark size=1.5pt]
              coordinates {(3-yr,0.27) (5-yr,0.16) (8-yr,0.11)};
            \addplot[barcolor1,thick,mark=*,mark size=1.5pt]
              coordinates {(3-yr,0.17) (5-yr,0.11) (8-yr,0.07)};
            \addplot[barcolor4,thick,mark=triangle*,mark size=2pt]
              coordinates {(3-yr,0.12) (5-yr,0.08) (8-yr,0.06)};
            \addplot[barcolor5,thick,mark=diamond*,mark size=2pt]
              coordinates {(3-yr,0.20) (5-yr,0.19) (8-yr,0.18)};
          \end{axis}
        \end{tikzpicture}

        \vspace{0.15cm}

        {\scriptsize
          \tikz[baseline=-0.5ex]{\draw[barcolor2,thick] (0,0.09cm) -- (0.35cm,0.09cm);
            \fill[barcolor2] (0.125cm,0.04cm) rectangle (0.225cm,0.14cm);}~8$\times$ B200
          \enspace
          \tikz[baseline=-0.5ex]{\draw[barcolor1,thick] (0,0.09cm) -- (0.35cm,0.09cm);
            \fill[barcolor1] (0.175cm,0.09cm) circle (0.05cm);}~8$\times$ H100
          \enspace
          \tikz[baseline=-0.5ex]{\draw[barcolor4,thick] (0,0.09cm) -- (0.35cm,0.09cm);
            \fill[barcolor4] (0.175cm,0.02cm) -- (0.24cm,0.09cm) -- (0.175cm,0.16cm) -- (0.11cm,0.09cm) -- cycle;}~8$\times$ A100
          \enspace
          \tikz[baseline=-0.5ex]{\draw[barcolor5,thick] (0,0.09cm) -- (0.35cm,0.09cm);
            \fill[barcolor5] (0.175cm,0.02cm) -- (0.24cm,0.09cm) -- (0.175cm,0.16cm) -- (0.11cm,0.09cm) -- cycle;}~128 V100
        }
        {\centering\caption{LLaMA 3.1-70B (Decode-heavy)}\label{fig:cost-70b}\par}

    \end{subfigure}

    \caption{Total cost of ownership (USD per million tokens) across
    four energy mixes and deployment lifetimes of 3, 5, and 8~years
    (data from \Cref{tab:cost-8b:prefill,tab:cost-70b-decode}).
    For the 8B model~(a), the 128-V100 \sysname{} is the cheapest
    option across all mixes.
    For the 70B model on decode-heavy workload~(b), V100 pods do not always
    achieve the lowest cost, but demonstrate superior performance in certain
    combinations of workload, model size, and energy mix.}
    \label{fig:cost-comparison}
\end{figure*}
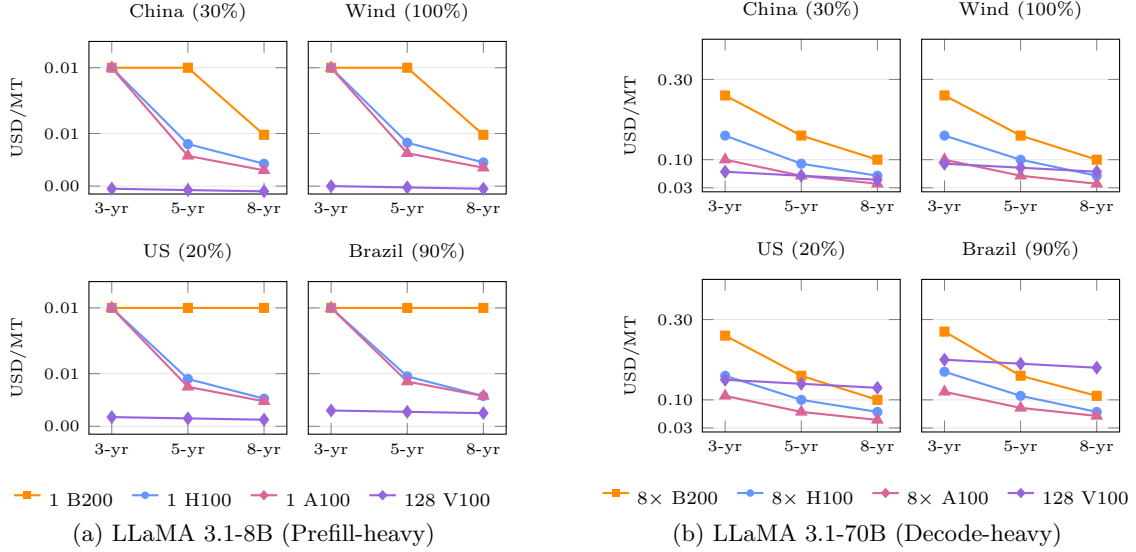

We also evaluate the combined capital and operational expenditures over a 3-year, 5-year, and 8-year deployment lifetime (Cost 3-year, Cost 5-year and Cost 8-year) in \Cref{tab:cost-8b:prefill} and \Cref{tab:cost-70b:prefill}; these costs are visualized in \Cref{fig:cost-comparison}. Despite the higher operational costs associated with V100-based systems, their reduced capital expenses make them a financially viable platform for deployment. Our analysis reveals that for an 8B model, when utilizing a carefully selected energy supply (China Energy Mix), a 128-V100 pod can be up to $6\times$ more cost-effective compared to H100 counterparts. The cost-effectiveness gap shortens with 70B models: under the China energy mix, the 128-V100 pod's 3-year total cost per token is approximately $1.5\times$ that of an H100 system (i.e., $67\%$ of H100's cost efficiency), compared to the $6\times$ advantage observed for 8B models. This reflects the higher per-token operational cost of older hardware at larger model scales. This demonstrates that, for end-of-life GPUs, like these V100-based systems, with adequate engineering and system co-development, still show some competitive advantage in deployment, primarily due to their significantly lower capital cost.
We then investigate the potential environmental impacts of running LLM inference with these V100-based systems. \Cref{tab:cost-8b:prefill} and \Cref{tab:cost-70b:prefill} show the carbon footprint ($C_{\text{total}}$, \Cref{eq:ctotal}) of different hardware, which combines operational carbon ($C_{\text{opt}}$, \Cref{eq:copt}) and embodied carbon ($C_{\text{em}}$, \Cref{eq:cem}). We consider a 5-year lifetime for all hardware when calculating the carbon footprint.
The lifetime carbon footprint of V100s can be perceived in different ways. One perspective is to consider that, by repurposing V100s for LLM serving, their lifespan is effectively extended, which we show as `Carbon Footprint (extended)' in both \Cref{tab:cost-8b:prefill} and \Cref{tab:cost-70b:prefill}. Alternatively, if we consider that these devices, once phased out of data centers, would have reached the end of their useful life---with their embodied carbon effectively zeroed since they are being reused instead of discarded---we can present this as `Carbon Footprint (zero embodied)'. This basically assumes that, without reuse, the devices would end up in landfills.
We also take into account the operational carbon emissions from different energy supplies and factor in the embodied carbon to calculate the carbon footprint per million token generation.
As seen in both \Cref{tab:cost-8b:prefill} and \Cref{tab:cost-70b:prefill}, our approach using second-hand GPUs exhibits a larger carbon footprint. For the 8B model, the carbon footprint could be approximately 3.7 times larger, and for the 70B model, this disparity increases to approximately 41 times due to the inherent increase in electricity usage of V100 deployment.

\begin{table}[!h]
\centering
\caption{\textbf{Prefill-heavy workload, LLaMA 3.1-8B.}, Running model serving on new and second-hand GPUs with LLaMA 3.1-8B. We measure Token Per Second (TPS) and actual Operational cost (USD per million Tokens) with a unit electricity price from the different energy mixes with Prefill-heavy tasks.
We also consider the combined capital and operational costs with a 3-year, 5-year, and an 8-year life-time.
We report carbon footprints, for V100s, we consider two values ``Carbon Footprint (extended)/Carbon Footprint (zero embodied)'', which are further explained in \Cref{sec:results}.
}
\label{tab:cost-8b:prefill}
\begin{tabular}{@{}cc|ccc|cc@{}}
\toprule
\multirow{2}{*}{\textbf{Metrics}}
& \multirow{2}{*}{\textbf{Units}}
& \textbf{B200}$^\dagger$
& \textbf{H100}
& \textbf{A100 80GB}
& \textbf{V100}
& \textbf{V100} \\
& & 1 GPU & 1 GPU & 1 GPU & 4-GPU pod & 128-GPU pod \\
\midrule
\multicolumn{7}{c}{Inference serving metrics}                    \\ \midrule
Capital cost        & USD             & 75K  & 18.5K  & 5.8K  & 240  & 7.68K  \\

TPS                  & tokens/s
& 62651  & 29343  & 12451  & 6998  & 223936  \\ \midrule
\multicolumn{7}{c}{$30\%$ renewable energy (China energy mix)}                           \\ \midrule
Operational cost       & USD/MT        & \textbf{0.0001} & 0.0002 & 0.0004 & 0.0005 & 0.0005 \\
Cost 3-year & USD/MT        & 0.01 & 0.01 & 0.01 & \textbf{0.0008} & \textbf{0.0008} \\
Cost 5-year & USD/MT        & 0.01 & 0.0042 & 0.0033 & \textbf{0.0007} & \textbf{0.0007} \\
Cost 8-year & USD/MT        & 0.0049 & 0.0027 & 0.0022 & \textbf{0.0006} & \textbf{0.0006} \\
Carbon footprint          & gCO\textsubscript{2}/MT          & \textbf{1.84} & 2.82 & 5.35 & 6.97/6.65* & 6.85/6.65* \\
\midrule
\multicolumn{7}{c}{$100\%$ renewable energy (China Wind Farm )}                           \\ \midrule
Operational cost       & USD/MT        & \textbf{0.0002} & 0.0003 & 0.0005 & 0.0007 & 0.0007 \\
Cost 3-year & USD/MT        & 0.01 & 0.01 & 0.01 & \textbf{0.0010} & \textbf{0.0010} \\
Cost 5-year & USD/MT        & 0.01 & 0.0043 & 0.0035 & \textbf{0.0009} & \textbf{0.0009} \\
Cost 8-year & USD/MT        & 0.0049 & 0.0028 & 0.0024 & \textbf{0.0008} & \textbf{0.0008} \\
Carbon footprint          & gCO\textsubscript{2}/MT          & \textbf{0.07} & 0.11 & 0.24 & 0.43/0.11* & 0.31/0.11* \\
\midrule
\multicolumn{7}{c}{$20\%$ renewable energy (US energy mix)}                           \\ \midrule
Operational cost       & USD/MT        & \textbf{0.0004} & 0.0006 & 0.0011 & 0.0014 & 0.0014 \\
Cost 3-year & USD/MT        & 0.01 & 0.01 & 0.01 & \textbf{0.0017} & \textbf{0.0017} \\
Cost 5-year & USD/MT        & 0.01 & 0.0046 & 0.0040 & \textbf{0.0016} & \textbf{0.0016} \\
Cost 8-year & USD/MT        & 0.01 & 0.0031 & 0.0029 & \textbf{0.0015} & \textbf{0.0015} \\
Carbon footprint          & gCO\textsubscript{2}/MT          & \textbf{1.15} & 1.77 & 3.37 & 4.43/4.12* & 4.31/4.12* \\
\midrule
\multicolumn{7}{c}{$90\%$ renewable energy (Brazil energy mix)}                           \\ \midrule
Operational cost       & USD/MT        & \textbf{0.0005} & 0.0008 & 0.0015 & 0.0019 & 0.0019 \\
Cost 3-year & USD/MT        & 0.01 & 0.01 & 0.01 & \textbf{0.0022} & \textbf{0.0022} \\
Cost 5-year & USD/MT        & 0.01 & 0.0048 & 0.0044 & \textbf{0.0021} & \textbf{0.0021} \\
Cost 8-year & USD/MT        & 0.01 & 0.0033 & 0.0033 & \textbf{0.0020} & \textbf{0.0020} \\
Carbon footprint          & gCO\textsubscript{2}/MT          & \textbf{0.34} & 0.53 & 1.02 & 1.42/1.11* & 1.31/1.11* \\
\bottomrule
\end{tabular}
\begin{tablenotes}
\footnotesize
\item[$\dagger$] B200 results were profiled with vLLM v0.17; H100, A100, and V100 benchmarks were profiled using vLLM v0.6.
\item[*] For V100 carbon footprint, values are reported as ``extended/zero embodied'' (see \Cref{sec:results}).
\end{tablenotes}
\end{table}

\begin{table}[!ht]
\centering
\caption{\textbf{Prefill-heavy workload, LLaMA 3.1-70B.} Running model serving on new and second-hand GPUs with LLaMA 3.1-70B. We measure Token Per Second (TPS) and actual Operational cost (USD per million Tokens) with a unit electricity price from the different energy mixes with Prefill-heavy tasks.
We also consider the combined capital and operational costs with a 3-year, 5-year, and an 8-year life-time.
We report carbon footprints, for V100s, we consider two values ``Carbon Footprint (extended)/Carbon Footprint (zero embodied)'', which are further explained in \Cref{sec:results}.
}
\label{tab:cost-70b:prefill}

\begin{tabular}{@{}cc|ccc|cc@{}}
\toprule
\multirow{2}{*}{\textbf{Metrics}}
& \multirow{2}{*}{\textbf{Units}}
& \textbf{B200}$^\dagger$
& \textbf{H100}
& \textbf{A100 80GB}
& \textbf{V100}
& \textbf{V100} \\
& & 8-GPU pod & 8-GPU pod & 8-GPU pod & 16-GPU pod & 128-GPU pod \\
\midrule
\multicolumn{7}{c}{Inference serving metrics}                    \\ \midrule
Capital cost        & USD             & 600K  & 148K  & 46.4K  & 960  & 7.68K  \\

TPS                  & tokens/s
& 37045  & 19885  & 8642  & 189  & 1512  \\ \midrule
\multicolumn{7}{c}{$30\%$ renewable energy (China energy mix)}                           \\ \midrule
Operational cost       & USD/MT        & \textbf{0.0017} & 0.0022 & 0.0041 & 0.07 & 0.07 \\
Cost 3-year & USD/MT        & 0.17 & 0.08 & \textbf{0.06} & 0.12 & 0.12 \\
Cost 5-year & USD/MT        & 0.10 & 0.05 & \textbf{0.04} & 0.10 & 0.10 \\
Cost 8-year & USD/MT        & 0.07 & 0.03 & \textbf{0.03} & 0.09 & 0.09 \\
Carbon footprint          & gCO\textsubscript{2}/MT          & \textbf{24.62} & 32.74 & 60.36 & 1014.91/985.40* & 1014.91/985.40* \\
\midrule
\multicolumn{7}{c}{$100\%$ renewable energy (China Wind Farm)}                           \\ \midrule
Operational cost       & USD/MT        & \textbf{0.0025} & 0.0034 & 0.01 & 0.10 & 0.10 \\
Cost 3-year & USD/MT        & 0.17 & 0.08 & \textbf{0.06} & 0.16 & 0.16 \\
Cost 5-year & USD/MT        & 0.11 & 0.05 & \textbf{0.04} & 0.13 & 0.13 \\
Cost 8-year & USD/MT        & 0.07 & 0.03 & \textbf{0.03} & 0.12 & 0.12 \\
Carbon footprint          & gCO\textsubscript{2}/MT          & \textbf{0.61} & 0.78 & 1.53 & 46.45/16.93* & 46.45/16.93* \\
\midrule
\multicolumn{7}{c}{$20\%$ renewable energy (US energy mix)}                           \\ \midrule
Operational cost       & USD/MT        & \textbf{0.01} & 0.01 & 0.01 & 0.20 & 0.20 \\
Cost 3-year & USD/MT        & 0.18 & 0.09 & \textbf{0.07} & 0.26 & 0.26 \\
Cost 5-year & USD/MT        & 0.11 & 0.05 & \textbf{0.05} & 0.24 & 0.24 \\
Cost 8-year & USD/MT        & 0.07 & 0.04 & \textbf{0.03} & 0.22 & 0.22 \\
Carbon footprint          & gCO\textsubscript{2}/MT          & \textbf{15.30} & 20.34 & 37.53 & 639.04/609.52* & 639.04/609.52* \\
\midrule
\multicolumn{7}{c}{$90\%$ renewable energy (Brazil energy mix)}                           \\ \midrule
Operational cost       & USD/MT        & \textbf{0.01} & 0.01 & 0.02 & 0.28 & 0.28 \\
Cost 3-year & USD/MT        & 0.18 & 0.09 & \textbf{0.07} & 0.33 & 0.33 \\
Cost 5-year & USD/MT        & 0.11 & 0.06 & \textbf{0.05} & 0.31 & 0.31 \\
Cost 8-year & USD/MT        & 0.07 & 0.04 & \textbf{0.04} & 0.30 & 0.30 \\
Carbon footprint          & gCO\textsubscript{2}/MT          & \textbf{4.27} & 5.66 & 10.50 & 194.14/164.62* & 194.14/164.62* \\
\bottomrule
\end{tabular}
\begin{tablenotes}
\footnotesize
\item[$\dagger$] B200 results were profiled with vLLM v0.17; H100, A100, and V100 benchmarks were profiled using vLLM v0.6.
\item[*] For V100 carbon footprint, values are reported as ``extended/zero embodied'' (see \Cref{sec:results}).
\end{tablenotes}
\end{table}

%
%

\section{Discussion and Limitations}\label{sec:discussion}

Our work demonstrates that second-hand GPUs can offer competitive performance and cost-effectiveness for LLM inference when paired with specialized software optimizations and appropriate deployment conditions. However, several important considerations and limitations warrant careful discussion.

\subsection{Hardware Reliability and Operational Challenges}

While we have implemented hardware binning and stress-testing procedures to mitigate reliability concerns (\Cref{sec:method}), second-hand hardware reliability remains an issue for production deployments. Unlike new hardware with full manufacturer warranties and predictable failure rates, second-hand devices carry inherent uncertainty about their operational history and remaining lifespan.

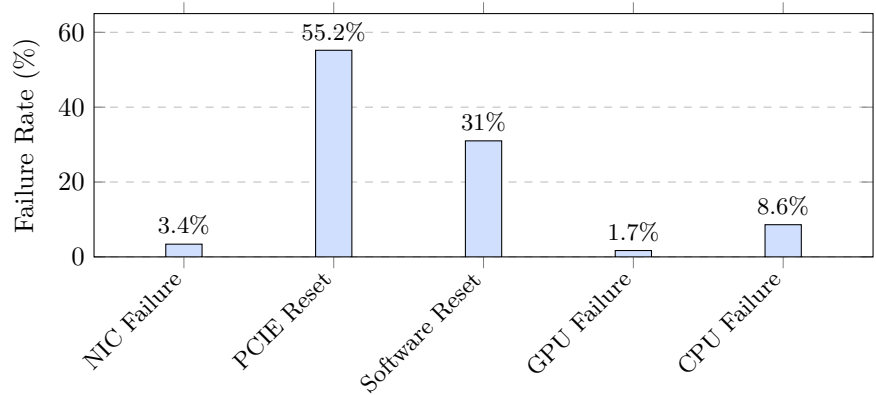
\begin{figure}[!t]
    \centering
    \definecolor{barcolor1}{RGB}{100, 150, 255}
    \begin{tikzpicture}
        \begin{axis}[
            ybar,
            bar width=0.48cm,
            width=0.72\textwidth,
            height=4.8cm,
            ylabel={Failure Rate (\%)},
            symbolic x coords={NIC Failure, PCIE Reset, Software Reset, GPU Failure, CPU Failure},
            xtick=data,
            x tick label style={rotate=45, anchor=east, font=\small},
            ymin=0,
            ymax=65,
            ymajorgrids=true,
            grid style=dashed,
            enlarge x limits=0.15,
            nodes near coords={\pgfmathprintnumber\pgfplotspointmeta\%},
            every node near coord/.append style={font=\small},
        ]
        \addplot[fill=barcolor1!30] coordinates {
            (NIC Failure, 3.4)
            (PCIE Reset, 55.2)
            (Software Reset, 31.0)
            (GPU Failure, 1.7)
            (CPU Failure, 8.6)
        };
        \end{axis}
    \end{tikzpicture}
    \caption{Hardware failure rate breakdown from $1$ year of actual operating deployment of our 128-GPU V100 pod. Out of all service impacting incidents, only 13.8\% were true hardware failures requiring device replacement (NIC, GPU, CPU), while the majority (86.2\%) were recoverable resets handled through software interventions.}
    \label{fig:hardware-failures}
\end{figure}

To empirically assess reliability in practice, we tracked all hardware incidents across our 128-GPU V100 cluster over one year of continuous operation. \Cref{fig:hardware-failures} presents the breakdown of failure types observed during this period. Across 58 total incidents, we found that true hardware failures requiring device replacement were relatively rare: only 8 cases (13.8\%) consisted of actual hardware failures---2 NIC failures, 1 GPU failure, and 5 CPU failures. The majority of incidents (50 cases, 86.2\%) were recoverable through software interventions: 32 PCIE resets and 18 software-reasoned resets. These recoverable failures typically required only service restarts or node reboots rather than physical hardware replacement. Notably, 50 total recoverable incidents over one year across 128 GPUs represents a relatively low failure rate.

This empirical data suggests that properly binned second-hand hardware can achieve acceptable reliability for production LLM serving. The 8 hardware failures over one year across 128 GPUs translate to approximately a 6.25\% annual device failure rate, which aligns with the $10\%$ additional capacity provisioning strategy we adopted based on prior reliability studies~\cite{ostrouchov2020gpu, kokolis2024revisiting}. The predominance of recoverable software-level failures indicates that robust cluster management software can effectively mask most reliability issues from end users.

However, organizations considering second-hand deployments must account for higher operational overhead in monitoring, maintenance, and replacement logistics compared to new hardware fleets. The lack of vendor support means organizations must develop in-house expertise for hardware diagnostics and repair, or accept higher device replacement rates. The economics of these trade-offs remain favorable when hardware capital costs dominate---as with V100 clusters at current market prices---but may shift as second-hand prices fluctuate or if failure rates exceed anticipated levels.

\subsection{Energy Infrastructure as a Critical Dependency}

Our results in \Cref{tab:cost-8b:prefill} and \Cref{tab:cost-70b:prefill} reveal a fundamental insight: the viability of second-hand GPU deployment is inextricably linked to energy infrastructure characteristics, specifically electricity cost and carbon intensity. This dependency has profound implications for the future of sustainable AI infrastructure and energy security.

The stark contrast in carbon footprints across different energy mixes --- ranging from sub-1 g/MT with $100\%$ renewable energy to approximately 7 g/MT with fossil-fuel dominated grids for 8B models --- demonstrates that hardware efficiency improvements alone cannot achieve sustainability goals. 
For second-hand deployments, such as the \sysname{}, to avoid significant negative environmental impacts, they must be strategically co-located with clean energy sources.

This presents both opportunities and challenges. Regions with abundant renewable energy capacity, such as areas with hydroelectric, geothermal, or wind resources, represent ideal deployment targets for second-hand AI infrastructure. Such deployments can achieve genuine environmental benefits by extending hardware lifecycles while leveraging low-carbon electricity. Conversely, deploying second-hand GPUs in regions dependent on fossil fuels may actually worsen the overall carbon footprint of AI services, despite the reduced embodied carbon from hardware reuse.

Moreover, the lower initial capital investments of second-hand hardware compared to new GPUs reduce the financial risk of deploying in regions with energy price volatility or regulatory uncertainty. 
Organizations can establish distributed inference capacity across multiple energy markets, reducing dependence on any single energy supplier or grid. This geographic diversification provides resilience against energy supply disruptions, price shocks, or policy changes that might affect concentrated data center deployments.

Looking forward, as AI inference workloads continue to grow, the energy demands will increasingly stress electrical grids. The lower energy efficiency of second-hand hardware exacerbates this challenge locally but enables strategic flexibility globally. Second-hand GPU clusters may find their most appropriate role in regions where renewable energy is abundant and inexpensive, effectively serving as a mechanism to monetize otherwise-curtailed renewable generation capacity while reducing pressure on constrained grid infrastructure in high-demand urban centers.

The importance of clean and cheap energy sources cannot be overstated for the future of AI inference. As inference demands scale with widespread AI deployment and compute-intensive paradigms like test-time scaling, the aggregate energy consumption will rival or exceed that of training. Without strategic coupling to low-carbon energy infrastructure, this growth will impose unsustainable environmental costs. Our findings suggest that future AI infrastructure planning must prioritize energy co-location: matching compute capacity deployment with renewable energy availability rather than treating energy as a commodity available anywhere at similar cost and carbon intensity. 

\subsection{Market Dynamics and Economic Sustainability}

Our economic analysis relies on current second-hand market prices, which reflect limited demand for older GPU models. However, this analysis contains an inherent paradox: if second-hand GPU deployment becomes widely adopted based on our demonstrated cost advantages, increased demand would likely drive up second-hand prices, potentially eroding the economic benefits we document.

The second-hand GPU market is currently characterized by excess supply, as hyperscale data centers regularly retire functional hardware to upgrade to newer generations. V100 prices have declined from approximately \$215 to \$50-\$60 over the past two years, reflecting a weak demand relative to the available supply. However, if our approach gains traction and organizations begin systematically acquiring second-hand GPUs for LLM serving, market dynamics would shift.

Beyond price dynamics, second-hand hardware offers important supply chain security advantages. The production of cutting-edge GPUs is concentrated among a few manufacturers and fabrication facilities, creating vulnerability to supply disruptions from geopolitical tensions, natural disasters, or capacity constraints. During recent GPU shortages, organizations faced procurement delays exceeding 6-12 months for new hardware, with some unable to secure allocations at all. Second-hand markets provide an alternative channel that is geographically distributed, less subject to export controls on frontier technology, and immediately accessible without multi-month lead times.

This supply chain diversification is particularly valuable for organizations in regions with limited access to new GPU supplies, whether due to vendor prioritization of hyperscale customers, export restrictions, or geopolitical factors. Second-hand hardware enables these organizations to participate in AI development and deployment without dependence on fragile supply chains for cutting-edge components. However, this supply chain independence comes with the trade-off of relying on hardware that major vendors no longer support, requiring organizations to develop alternative support channels and accept greater self-reliance in maintenance and repair.

\subsection{Workload Specificity: Inference versus Training}

A critical limitation of our approach is its primary applicability to inference workloads rather than training. This specialization stems from fundamental differences in computational characteristics between these workloads.

LLM training requires:
\begin{enumerate}
    \item High inter-GPU bandwidth for gradient synchronization across data-parallel and model-parallel dimensions
    \item Large batch sizes to maintain computational efficiency and training stability
    \item Precision in numerical computations (often requiring FP32 or mixed-precision training with careful gradient scaling)
    \item Extended continuous operation over weeks or months, amplifying the impact of any hardware reliability issues
\end{enumerate}

The limited NVLink bandwidth (300 GB/s for V100 versus 900 GB/s for H100) would bottleneck gradient synchronization across devices. Furthermore, training's reliability requirements are more stringent---a single device failure during a multi-week training run can require checkpoint rollback and substantial re-computation, making the higher failure rates of second-hand hardware particularly costly for training workloads.
However, the computational economics of LLM deployment are increasingly dominated by inference rather than training. A model is trained once but serves millions or billions of queries over its deployment lifetime, and this is where this second-hand GPU inference paradigm holds value.

\section{Conclusion}\label{sec:conclusion}

This paper demonstrates that second-hand GPUs, when paired with specialized software optimizations and deployed under appropriate conditions, offer a viable pathway for expanding AI inference capacity. We built a 128-GPU V100 cluster from entirely second-hand components, deployed it in production for one year, and developed a pipeline-first parallelization strategy that enables competitive LLM serving performance despite the hardware limitations of older devices.
Our empirical analysis reveals that the viability of GPU afterlife depends critically on three factors: workload characteristics, energy infrastructure, and market conditions. For inference workloads, which increasingly dominate AI's computational footprint, second-hand hardware can achieve throughput competitive with cutting-edge systems at a fraction of the capital cost (\$22K vs. \$600K). However, the environmental benefits materialize only when deployments are strategically co-located with low-carbon energy sources; otherwise, higher operational emissions---approximately $4\times$ for 8B and over $40\times$ for 70B models under grid-average carbon intensity---can negate the reduced embodied carbon.

\clearpage
\appendix
\crefalias{section}{appendix} 

\begin{appendices}

\section{Regional Energy Supply Characteristics}
\label{app:energy-grid}

We derive all the energy prices from the industrial sector energy price offered by various statistics sources. For the USA, we refer to the EIA statistics~\cite{eia_industrial_price_2025}. For China, we use the industrial electricity cost reported from CEIC~\cite{ceic_china_industrial_price_2025}. For Brazil the electricity cost statistics are derived from~\cite{globalpetrolprices_brazil}. Carbon intensity values are obtained from regional grid emission factors for 2023~\cite{ember2026carbon}, accounting for the energy mix composition of each region. Renewable energy penetration percentages are derived from~\cite{ember2026carbon} from 2023 as well. These data sources are collected from Ember with major processing of Our World in Data.
Table~\ref{tab:energy-supply} presents the comprehensive regional energy grid characteristics used in our analysis, including renewable energy penetration, carbon intensity, and industrial electricity prices.

\begin{table}[!h]
\centering
\caption{Regional energy supply characteristics}
\label{tab:energy-supply}
\begin{tabular}{lccc}
\hline
\textbf{Region} & \textbf{Renewable} & \textbf{Carbon Intensity} & \textbf{Energy Price} \\
 & \textbf{(\%)} & \textbf{(gCO$_2$/kWh)} & \textbf{(USD/kWh)} \\
\hline
China (Non-renewable) & $\sim$30 & 584.0 & 0.0556 \\
United States & $\sim$20 & 393.0 & 0.12 \\
Brazil & $\sim$90 & 97.0 & 0.125 \\
China (Renewable) & $\sim$100 & $\sim$10.0 & 0.0556 \\
\hline
\end{tabular}
\end{table}

The regional variations reflect different energy infrastructure and policy environments:
\begin{itemize}
    \item \textbf{China (Non-renewable)}: Represents deployment in regions with coal-dominated grids (30\% renewable penetration), resulting in high carbon intensity
    \item \textbf{United States}: Mixed grid with moderate renewable penetration (20\%) from various sources including natural gas, nuclear, and renewables
    \item \textbf{Brazil}: High renewable content (90\%) primarily from hydroelectric sources, resulting in low carbon intensity despite higher electricity costs
    \item \textbf{China (Renewable only)}: Represents deployment co-located with renewable energy sources (100\% renewable), achieving minimal operational carbon emissions
\end{itemize}

\section{GPU Price Calculation}

The hardware cost estimates utilized in this analysis are reported in United States Dollars (USD) and derived from a combination of direct vendor quotations and secondary market valuations. Specifically, the pricing for the NVIDIA B200 is based on a full-system quotation of \$600,000 obtained from Scan.co.uk. To calculate the individual per-unit GPU cost, this aggregate system price---which is inclusive of all requisite peripheral components and networking hardware---is divided by eight. Conversely, the cost estimates for the NVIDIA H100 and A100 accelerators are sourced from the secondary market; consequently, these figures are subject to inherent market volatility and may exhibit temporal fluctuations. Individual NVIDIA V100 prices are similarly derived from secondary market data. However, it should be noted that the cost for the 128-GPU V100 pod configuration is reported as a comprehensive total system cost, encompassing both the computational units and all necessary peripheral infrastructure.

\section{Workload Characteristics}
\label{appendix:workload}

To evaluate model performance across varied inference profiles, we utilized three distinct workload configurations (summarized in Table \ref{tab:token_workloads}). All workloads utilize a baseline 17-token system prompt.

\begin{table}[htbp]
\centering
\caption{Token distributions across different LLM workloads}
\label{tab:token_workloads}
\begin{tabular}{lcc}
\toprule
\textbf{Workload} & \textbf{Input Tokens} & \textbf{Output Tokens (max)} \\ \midrule
Prefill-heavy & 1,259 & 128 \\
Decode-heavy & 199 & 1,024 \\
SharedGPT & $\sim$262 mean / $\sim$94 median (variable) & 180 \\ \bottomrule
\end{tabular}
\end{table}

\begin{itemize}
    \item \textbf{Prefill-Heavy:} Evaluates prompt processing bottlenecks using a large input context (1,259 tokens, dominated by tool definitions) and a strictly capped 128-token generation.
    \item \textbf{Decode-Heavy:} Assesses auto-regressive generation efficiency using a concise input (199 tokens) and an extended generation window capped at 1,024 tokens.
    \item \textbf{ShareGPT:} Simulates real-world variability using 500 sampled conversations from the ShareGPT dataset (seed=42), chunked to 2,048 tokens per FastChat protocols, with a 180-token maximum output.
\end{itemize}

\section{Additional Experiment Result}

\subsection{Decode Heavy Experiment Results}

We present the cost and carbon footprint results under the Decode-heavy workload, which uses a concise 199-token input and an extended 1,024-token generation window (see \Cref{appendix:workload}). This workload emphasizes auto-regressive decoding throughput and is representative of conversational AI applications where output length dominates.

\begin{table}[!ht]
\centering
\caption{\textbf{Decode-heavy workload, LLaMA3.1-8B.} Running model serving on new and second-hand GPUs with LLaMA 3.1-8B. We measure Token Per Second (TPS) and actual Operational cost (USD per million Tokens) with a unit electricity price from the different energy mixes.
We also consider the combined capital and operational costs with a 3-year, 5-year, and an 8-year life-time.
We report carbon footprints, for V100s, we consider two values ``Carbon Footprint (extended)/Carbon Footprint (zero embodied)'', which are further explained in \Cref{sec:results}.
}
\label{tab:cost-8b-decode}
\begin{tabular}{@{}cc|ccc|cc@{}}
\toprule
\multirow{2}{*}{\textbf{Metrics}}
& \multirow{2}{*}{\textbf{Units}}
& \textbf{B200}$^\dagger$
& \textbf{H100}
& \textbf{A100 80GB}
& \textbf{V100}
& \textbf{V100} \\
& & 1 GPU & 1 GPU & 1 GPU & 4-GPU pod & 128-GPU pod \\
\midrule
\multicolumn{7}{c}{Inference serving metrics}                    \\ \midrule
Capital cost        & USD             & 75K  & 18.5K  & 5.8K  & 240  & 7.68K  \\

TPS                  & tokens/s
& \textbf{26037}  & 11506  & 5919  & 254  & 8128  \\ \midrule
\multicolumn{7}{c}{$30\%$ renewable energy (China energy mix)}                           \\ \midrule
Operational cost       & USD/MT        & \textbf{0.0003} & 0.0005 & 0.0008 & 0.01 & 0.01 \\
Cost 3-year & USD/MT        & 0.03 & 0.02 & \textbf{0.01} & 0.02 & 0.02 \\
Cost 5-year & USD/MT        & 0.02 & 0.01 & \textbf{0.01} & 0.02 & 0.02 \\
Cost 8-year & USD/MT        & 0.01 & 0.01 & \textbf{0.0046} & 0.02 & 0.02 \\
Carbon footprint          & gCO\textsubscript{2}/MT          & \textbf{4.43} & 7.19 & 11.25 & 191.92/183.31* & 188.80/183.31* \\
\midrule
\multicolumn{7}{c}{$100\%$ renewable energy (China Wind Farm)}                           \\ \midrule
Operational cost       & USD/MT        & \textbf{0.0004} & 0.0007 & 0.0011 & 0.02 & 0.02 \\
Cost 3-year & USD/MT        & 0.03 & 0.02 & \textbf{0.01} & 0.03 & 0.03 \\
Cost 5-year & USD/MT        & 0.02 & 0.01 & \textbf{0.01} & 0.02 & 0.02 \\
Cost 8-year & USD/MT        & 0.01 & 0.01 & \textbf{0.01} & 0.02 & 0.02 \\
Carbon footprint          & gCO\textsubscript{2}/MT          & \textbf{0.16} & 0.29 & 0.51 & 11.76/3.15* & 8.64/3.15* \\
\midrule
\multicolumn{7}{c}{$20\%$ renewable energy (US energy mix)}                           \\ \midrule
Operational cost       & USD/MT        & \textbf{0.0009} & 0.0014 & 0.0023 & 0.04 & 0.04 \\
Cost 3-year & USD/MT        & 0.03 & 0.02 & \textbf{0.01} & 0.05 & 0.05 \\
Cost 5-year & USD/MT        & 0.02 & 0.01 & \textbf{0.01} & 0.04 & 0.04 \\
Cost 8-year & USD/MT        & 0.01 & 0.01 & \textbf{0.01} & 0.04 & 0.04 \\
Carbon footprint          & gCO\textsubscript{2}/MT          & \textbf{2.77} & 4.51 & 7.08 & 122.00/113.39* & 118.88/113.39* \\
\midrule
\multicolumn{7}{c}{$90\%$ renewable energy (Brazil energy mix)}                           \\ \midrule
Operational cost       & USD/MT        & \textbf{0.0012} & 0.0020 & 0.0031 & 0.05 & 0.05 \\
Cost 3-year & USD/MT        & 0.03 & 0.02 & \textbf{0.01} & 0.06 & 0.06 \\
Cost 5-year & USD/MT        & 0.02 & 0.01 & \textbf{0.01} & 0.06 & 0.06 \\
Cost 8-year & USD/MT        & 0.01 & 0.01 & \textbf{0.01} & 0.06 & 0.06 \\
Carbon footprint          & gCO\textsubscript{2}/MT          & \textbf{0.81} & 1.34 & 2.15 & 39.23/30.62* & 36.11/30.62* \\
\bottomrule
\end{tabular}

\begin{tablenotes}
\footnotesize
\item[$\dagger$] B200 results were profiled with vLLM v0.17; H100, A100, and V100 benchmarks were profiled using vLLM v0.6.
\item[*] For V100 carbon footprint, values are reported as ``extended/zero embodied'' (see \Cref{sec:results}).
\end{tablenotes}
\end{table}

\newpage

\begin{table}[!htbp]
\centering
\caption{\textbf{Decode-heavy workload, LLaMA 3.1-70B.} Running model serving on new and second-hand GPUs with LLaMA 3.1-70B. We measure Token Per Second (TPS) and actual Operational cost (USD per million Tokens) with a unit electricity price from the different energy mixes.
We also consider the combined capital and operational costs with a 3-year, 5-year, and an 8-year life-time.
We report carbon footprints, for V100s, we consider two values ``Carbon Footprint (extended)/Carbon Footprint (zero embodied)'', which are further explained in \Cref{sec:results}.
}
\label{tab:cost-70b-decode}

\resizebox{\textwidth}{!}{%
\begin{tabular}{@{}cc|ccc|cc@{}}
\toprule
\multirow{2}{*}{\textbf{Metrics}}
& \multirow{2}{*}{\textbf{Units}}
& \textbf{B200}$^\dagger$
& \textbf{H100}
& \textbf{A100 80GB}
& \textbf{V100}
& \textbf{V100} \\
& & 8-GPU pod & 8-GPU pod & 8-GPU pod & 16-GPU pod & 128-GPU pod \\
\midrule
\multicolumn{7}{c}{Inference serving metrics}                    \\ \midrule
Capital cost        & USD             & 600K  & 148K  & 46.4K  & 960  & 7.68K  \\

TPS                  & tokens/s
& \textbf{24695}  & 10372  & 5366  & 317  & 2536  \\ \midrule
\multicolumn{7}{c}{$30\%$ renewable energy (China energy mix)}                           \\ \midrule
Operational cost       & USD/MT        & \textbf{0.0025} & 0.0043 & 0.01 & 0.04 & 0.04 \\
Cost 3-year & USD/MT        & 0.26 & 0.16 & 0.10 & \textbf{0.07} & \textbf{0.07} \\
Cost 5-year & USD/MT        & 0.16 & 0.09 & 0.06 & \textbf{0.06} & \textbf{0.06} \\
Cost 8-year & USD/MT        & 0.10 & 0.06 & \textbf{0.04} & 0.05 & 0.05 \\
Carbon footprint          & gCO\textsubscript{2}/MT          & \textbf{36.94} & 62.78 & 97.22 & 605.11/587.51* & 605.11/587.51* \\
\midrule
\multicolumn{7}{c}{$100\%$ renewable energy (China Wind Farm)}                           \\ \midrule
Operational cost       & USD/MT        & \textbf{0.0038} & 0.01 & 0.01 & 0.06 & 0.06 \\
Cost 3-year & USD/MT        & 0.26 & 0.16 & 0.10 & \textbf{0.09} & \textbf{0.09} \\
Cost 5-year & USD/MT        & 0.16 & 0.10 & \textbf{0.06} & 0.08 & 0.08 \\
Cost 8-year & USD/MT        & 0.10 & 0.06 & \textbf{0.04} & 0.07 & 0.07 \\
Carbon footprint          & gCO\textsubscript{2}/MT          & \textbf{0.91} & 1.50 & 2.46 & 27.69/10.09* & 27.69/10.09* \\
\midrule
\multicolumn{7}{c}{$20\%$ renewable energy (US energy mix)}                           \\ \midrule
Operational cost       & USD/MT        & \textbf{0.01} & 0.01 & 0.02 & 0.12 & 0.12 \\
Cost 3-year & USD/MT        & 0.26 & 0.16 & \textbf{0.11} & 0.15 & 0.15 \\
Cost 5-year & USD/MT        & 0.16 & 0.10 & \textbf{0.07} & 0.14 & 0.14 \\
Cost 8-year & USD/MT        & 0.10 & 0.07 & \textbf{0.05} & 0.13 & 0.13 \\
Carbon footprint          & gCO\textsubscript{2}/MT          & \textbf{22.96} & 38.99 & 60.44 & 381.01/363.41* & 381.01/363.41* \\
\midrule
\multicolumn{7}{c}{$90\%$ renewable energy (Brazil energy mix)}                           \\ \midrule
Operational cost       & USD/MT        & \textbf{0.01} & 0.02 & 0.03 & 0.17 & 0.17 \\
Cost 3-year & USD/MT        & 0.27 & 0.17 & \textbf{0.12} & 0.20 & 0.20 \\
Cost 5-year & USD/MT        & 0.16 & 0.11 & \textbf{0.08} & 0.19 & 0.19 \\
Cost 8-year & USD/MT        & 0.11 & 0.07 & \textbf{0.06} & 0.18 & 0.18 \\
Carbon footprint          & gCO\textsubscript{2}/MT          & \textbf{6.40} & 10.84 & 16.91 & 115.75/98.15* & 115.75/98.15* \\
\bottomrule
\end{tabular}
}

\begin{tablenotes}
\footnotesize
\item[$\dagger$] B200 results were profiled with vLLM v0.17; H100, A100, and V100 benchmarks were profiled using vLLM v0.6.
\item[*] For V100 carbon footprint, values are reported as ``extended/zero embodied'' (see \Cref{sec:results}).
\end{tablenotes}
\end{table}

\newpage

\subsection{ShareGPT Experiment Results}

We present the cost and carbon footprint results under the ShareGPT workload, which samples 500 real-world conversations with variable input lengths ($\sim$262 mean tokens) and a 180-token maximum output (see \Cref{appendix:workload}). This workload captures realistic conversational variability encountered in production chatbot deployments.

\begin{table}[!ht]
\centering
\caption{\textbf{ShareGPT workload, LLaMA 3.1-8B.} Running model serving on new and second-hand GPUs with LLaMA 3.1-8B. We measure Token Per Second (TPS) and actual Operational cost (USD per million Tokens) with a unit electricity price from the different energy mixes.
We also consider the combined capital and operational costs with a 3-year, 5-year, and an 8-year life-time.
We report carbon footprints, for V100s, we consider two values ``Carbon Footprint (extended)/Carbon Footprint (zero embodied)'', which are further explained in \Cref{sec:results}.
}
\label{tab:cost-8b-sharegpt}
\begin{tabular}{@{}cc|ccc|cc@{}}
\toprule
\multirow{2}{*}{\textbf{Metrics}}
& \multirow{2}{*}{\textbf{Units}}
& \textbf{B200}$^\dagger$
& \textbf{H100}
& \textbf{A100 80GB}
& \textbf{V100}
& \textbf{V100} \\
& & 1 GPU & 1 GPU & 1 GPU & 4-GPU pod & 128-GPU pod \\
\midrule
\multicolumn{7}{c}{Inference serving metrics}                    \\ \midrule
Capital cost        & USD             & 75K  & 18.5K  & 5.8K  & 240  & 7.68K  \\

TPS                  & tokens/s
& \textbf{43911}  & 9312  & 4551  & 672  & 21504  \\ \midrule
\multicolumn{7}{c}{$30\%$ renewable energy (China energy mix)}                           \\ \midrule
Operational cost       & USD/MT        & \textbf{0.0002} & 0.0006 & 0.0010 & 0.0048 & 0.0048 \\
Cost 3-year & USD/MT        & 0.02 & 0.02 & 0.01 & \textbf{0.01} & \textbf{0.01} \\
Cost 5-year & USD/MT        & 0.01 & 0.01 & 0.01 & \textbf{0.01} & \textbf{0.01} \\
Cost 8-year & USD/MT        & 0.01 & 0.01 & \textbf{0.01} & 0.01 & 0.01 \\
Carbon footprint          & gCO\textsubscript{2}/MT          & \textbf{2.63} & 8.89 & 14.63 & 72.54/69.29* & 71.36/69.29* \\
\midrule
\multicolumn{7}{c}{$100\%$ renewable energy (China Wind Farm)}                           \\ \midrule
Operational cost       & USD/MT        & \textbf{0.0003} & 0.0009 & 0.0015 & 0.01 & 0.01 \\
Cost 3-year & USD/MT        & 0.02 & 0.02 & 0.01 & \textbf{0.01} & \textbf{0.01} \\
Cost 5-year & USD/MT        & 0.01 & 0.01 & 0.01 & \textbf{0.01} & \textbf{0.01} \\
Cost 8-year & USD/MT        & 0.01 & 0.01 & \textbf{0.01} & 0.01 & 0.01 \\
Carbon footprint          & gCO\textsubscript{2}/MT          & \textbf{0.10} & 0.36 & 0.67 & 4.44/1.19* & 3.27/1.19* \\
\midrule
\multicolumn{7}{c}{$20\%$ renewable energy (US energy mix)}                           \\ \midrule
Operational cost       & USD/MT        & \textbf{0.0005} & 0.0018 & 0.0029 & 0.01 & 0.01 \\
Cost 3-year & USD/MT        & 0.02 & 0.02 & \textbf{0.02} & 0.02 & 0.02 \\
Cost 5-year & USD/MT        & 0.01 & 0.01 & \textbf{0.01} & 0.02 & 0.02 \\
Cost 8-year & USD/MT        & \textbf{0.01} & 0.01 & 0.01 & 0.02 & 0.02 \\
Carbon footprint          & gCO\textsubscript{2}/MT          & \textbf{1.65} & 5.58 & 9.21 & 46.11/42.86* & 44.93/42.86* \\
\midrule
\multicolumn{7}{c}{$90\%$ renewable energy (Brazil energy mix)}                           \\ \midrule
Operational cost       & USD/MT        & \textbf{0.0007} & 0.0025 & 0.0040 & 0.02 & 0.02 \\
Cost 3-year & USD/MT        & 0.02 & 0.02 & \textbf{0.02} & 0.02 & 0.02 \\
Cost 5-year & USD/MT        & \textbf{0.01} & 0.02 & 0.01 & 0.02 & 0.02 \\
Cost 8-year & USD/MT        & \textbf{0.01} & 0.01 & 0.01 & 0.02 & 0.02 \\
Carbon footprint          & gCO\textsubscript{2}/MT          & \textbf{0.48} & 1.66 & 2.80 & 14.83/11.57* & 13.65/11.57* \\
\bottomrule
\end{tabular}

\begin{tablenotes}
\footnotesize
\item[$\dagger$] B200 results were profiled with vLLM v0.17; H100, A100, and V100 benchmarks were profiled using vLLM v0.6.
\item[*] For V100 carbon footprint, values are reported as ``extended/zero embodied'' (see \Cref{sec:results}).
\end{tablenotes}
\end{table}

\newpage

\begin{table}[!htbp]
\centering
\caption{\textbf{ShareGPT workload, LLaMA 3.1-70B.} Running model serving on new and second-hand GPUs with LLaMA 3.1-70B. We measure Token Per Second (TPS) and actual Operational cost (USD per million Tokens) with a unit electricity price from the different energy mixes.
We also consider the combined capital and operational costs with a 3-year, 5-year, and an 8-year life-time.
We report carbon footprints, for V100s, we consider two values ``Carbon Footprint (extended)/Carbon Footprint (zero embodied)", which are further explained in \Cref{sec:results}.
}
\label{tab:cost-70b-sharegpt}
\begin{tabular}{@{}cc|ccc|cc@{}}
\toprule
\multirow{2}{*}{\textbf{Metrics}}
& \multirow{2}{*}{\textbf{Units}}
& \textbf{B200}$^\dagger$
& \textbf{H100}
& \textbf{A100 80GB}
& \textbf{V100}
& \textbf{V100} \\
& & 8-GPU pod & 8-GPU pod & 8-GPU pod & 16-GPU pod & 128-GPU pod \\
\midrule
\multicolumn{7}{c}{Inference serving metrics}                    \\ \midrule
Capital cost        & USD             & 600K  & 148K  & 46.4K  & 960  & 7.68K  \\

TPS                  & tokens/s
& \textbf{41857}  & 5265  & 3402  & 440  & 3520  \\ \midrule
\multicolumn{7}{c}{$30\%$ renewable energy (China energy mix)}                           \\ \midrule
Operational cost       & USD/MT        & \textbf{0.0015} & 0.01 & 0.01 & 0.03 & 0.03 \\
Cost 3-year & USD/MT        & 0.15 & 0.31 & 0.15 & \textbf{0.05} & \textbf{0.05} \\
Cost 5-year & USD/MT        & 0.09 & 0.19 & 0.10 & \textbf{0.04} & \textbf{0.04} \\
Cost 8-year & USD/MT        & 0.06 & 0.12 & 0.06 & \textbf{0.04} & \textbf{0.04} \\
Carbon footprint          & gCO\textsubscript{2}/MT          & \textbf{21.79} & 123.67 & 153.34 & 435.95/423.27* & 435.95/423.27* \\
\midrule
\multicolumn{7}{c}{$100\%$ renewable energy (China Wind Farm)}                           \\ \midrule
Operational cost       & USD/MT        & \textbf{0.0022} & 0.01 & 0.02 & 0.04 & 0.04 \\
Cost 3-year & USD/MT        & 0.15 & 0.31 & 0.16 & \textbf{0.07} & \textbf{0.07} \\
Cost 5-year & USD/MT        & 0.09 & 0.19 & 0.10 & \textbf{0.06} & \textbf{0.06} \\
Cost 8-year & USD/MT        & 0.06 & 0.12 & 0.07 & \textbf{0.05} & \textbf{0.05} \\
Carbon footprint          & gCO\textsubscript{2}/MT          & \textbf{0.54} & 2.95 & 3.88 & 19.95/7.27* & 19.95/7.27* \\
\midrule
\multicolumn{7}{c}{$20\%$ renewable energy (US energy mix)}                           \\ \midrule
Operational cost       & USD/MT        & \textbf{0.0045} & 0.03 & 0.03 & 0.09 & 0.09 \\
Cost 3-year & USD/MT        & 0.16 & 0.32 & 0.18 & \textbf{0.11} & \textbf{0.11} \\
Cost 5-year & USD/MT        & \textbf{0.10} & 0.20 & 0.12 & 0.10 & 0.10 \\
Cost 8-year & USD/MT        & \textbf{0.06} & 0.14 & 0.09 & 0.10 & 0.10 \\
Carbon footprint          & gCO\textsubscript{2}/MT          & \textbf{13.54} & 76.82 & 95.33 & 274.50/261.82* & 274.50/261.82* \\
\midrule
\multicolumn{7}{c}{$90\%$ renewable energy (Brazil energy mix)}                           \\ \midrule
Operational cost       & USD/MT        & \textbf{0.01} & 0.03 & 0.04 & 0.12 & 0.12 \\
Cost 3-year & USD/MT        & 0.16 & 0.33 & 0.19 & \textbf{0.14} & \textbf{0.14} \\
Cost 5-year & USD/MT        & \textbf{0.10} & 0.21 & 0.13 & 0.13 & 0.13 \\
Cost 8-year & USD/MT        & \textbf{0.06} & 0.15 & 0.10 & 0.13 & 0.13 \\
Carbon footprint          & gCO\textsubscript{2}/MT          & \textbf{3.78} & 21.36 & 26.68 & 83.39/70.71* & 83.39/70.71* \\
\bottomrule
\end{tabular}

\begin{tablenotes}
\footnotesize
\item[$\dagger$] B200 results were profiled with vLLM v0.17; H100, A100, and V100 benchmarks were profiled using vLLM v0.6.
\item[*] For V100 carbon footprint, values are reported as ``extended/zero embodied'' (see \Cref{sec:results}).
\end{tablenotes}
\end{table}

\newpage

\section{Carbon Calculation}

We attribute the LLM carbon emissions at the model inference stage to two main
sources: (1)~\textbf{operational carbon} ($C_{\text{opt}}$), which is related
to the operational energy consumption of running LLM inferences, and
(2)~\textbf{embodied carbon} ($C_{\text{em}}$), which arises from the
production of hardware required to deliver the LLM inference service. The total
attributed LLM carbon emission is:
\begin{equation}\label{eq:ctotal}
  C_{\text{total}} = C_{\text{opt}} + C_{\text{em}}.
\end{equation}

\subsection{Operational carbon estimation}

Operational carbon ($C_{\text{opt}}$) is determined by the run-time energy
consumption of the underlying data center and the carbon intensity of its
energy supply:
\begin{equation}\label{eq:copt}
  C_{\text{opt}} = E_{\text{DC}} \times I,
\end{equation}
where $E_{\text{DC}}$ represents the total energy consumed by the data center
--- including energy consumed directly by the servers, cooling systems, and
other infrastructure components --- and $I$ (in gCO$_2$/kWh) is the carbon
intensity of the energy source.

\paragraph{Data center energy.}
We consider that all components of the server operate in the same data center.
Consequently, we model their operational energy consumption using:
\begin{equation}\label{eq:edc}
  E_{\text{DC}} = E_{\text{server}} \times \text{PUE},
\end{equation}
where PUE\footnote{Power Usage Effectiveness} captures the overhead from
cooling and other infrastructure. We adopt a PUE value of $1.2$ for all
hardware components, averaged across configurations reported in the literature:
high-efficiency setups described by
Faiz~\textit{et al.}~\cite{faiz2023llmcarbon} and high-density configurations
noted by Heydari~\textit{et al.}~\cite{heydari2022power}.

\paragraph{Carbon intensity and energy supply.}
The carbon intensity $I$ varies depending on the energy source. We consider
four representative energy mixes following data reported
in~\cite{owid-energy-mix}, summarised in \Cref{tab:energy-carbon-summary}:

\begin{table}[!h]
\centering
\caption{Energy supply configurations used in the carbon analysis. Carbon
intensity values are obtained from~\cite{owid_co2_per_unit_energy_gcb_2025}
and~\cite{uk_parliament_post_2006}. Energy prices are from regional industrial
tariffs. Full details are available in \Cref{tab:energy-supply}.}
\label{tab:energy-carbon-summary}
\begin{tabular}{lccc}
\toprule
\textbf{Energy Supply} & \textbf{Renewable} & \textbf{$I$ (gCO$_2$/kWh)} & \textbf{Price (USD/kWh)} \\
\midrule
US energy mix          & $\sim$20\%  & 393   & 0.12   \\
China energy mix       & $\sim$30\%  & 584   & 0.0556 \\
Brazil energy mix      & $\sim$90\%  & 97    & 0.125  \\
China (renewable only) & $\sim$100\% & $\sim$10 & 0.0556 \\
\bottomrule
\end{tabular}
\end{table}

\noindent These four energy configurations represent a spectrum from
fossil-fuel-dominated grids (US at $\sim$20\% renewable) to fully renewable
supplies (direct supply from a wind farm or hydroelectric station in China).
For even the completely renewable energy supply, a residual carbon intensity of
approximately $10$\,g/kWh remains due to lifecycle emissions from renewable
infrastructure~\cite{uk_parliament_post_2006}. It is also worth noting that
the energy supply affects the operational cost reported in
\Cref{tab:cost-8b:prefill,tab:cost-70b:prefill}.

\subsection{Embodied carbon estimation}

Embodied carbon ($C_{\text{em}}$) quantifies the manufacturing-stage carbon emissions
of the hardware used to deliver LLM inference. We decompose $C_{\text{em}}$ into three
additive terms:
\begin{equation}\label{eq:cem}
  C_{em} = C_{\text{die}} + C_{\text{VRAM}} + C_{\text{periph}},
\end{equation}
where $C_{\text{die}}$ is the logic-die embodied carbon, $C_{\text{VRAM}}$ is
the memory (VRAM) embodied carbon, and $C_{\text{periph}}$ is the peripheral
(server-level) embodied carbon.

\paragraph{Logic-die embodied carbon.}
We estimate $C_{\text{die}}$ from the GPU silicon area and the carbon per area
(CPA) of its manufacturing process node:
\begin{equation}\label{eq:cdie}
  C_{\text{die}} = A_{\text{die}} \times \text{CPA},
\end{equation}
where $A_{\text{die}}$ (in cm$^2$) is the die area obtained from technical
whitepapers~\cite{nvidia_volta_wp2017, nvidia_ampere_wp2020,
nvidia_hopper_wp2022}, and CPA (in kg\,CO$_2$e/cm$^2$) is sourced from the
methodologies of Li~\textit{et al.}~\cite{li2024towards} and
Faiz~\textit{et al.}~\cite{faiz2023llmcarbon}.

\paragraph{VRAM embodied carbon.}
We compute $C_{\text{VRAM}}$ from the HBM capacity and a unit embodied-carbon
factor:
\begin{equation}\label{eq:cvram}
  C_{\text{VRAM}} = S_{\text{VRAM}} \times U_{\text{VRAM}},
\end{equation}
where $S_{\text{VRAM}}$ (in GB) is the total HBM size and $U_{\text{VRAM}}$
(in kg\,CO$_2$e/GB) is the unit embodied carbon obtained from
Li~\textit{et al.}~\cite{li2024towards}.

\paragraph{Peripheral embodied carbon.}
Each GPU hosting server shares a common set of components (CPU, RAM, chassis,
etc.). We adopt a server-level peripheral cost of $C_{\text{periph}} = 150$\,kg
CO$_2$e per server, following the measurements of
Faiz~\textit{et al.}~\cite{faiz2023llmcarbon}.

\paragraph{Annualized embodied carbon.}
To compare devices across different deployment durations, we amortize the total
embodied carbon over the expected lifespan $T$ (in years):
\begin{equation}\label{eq:annual}
  C_{em}^{\text{annual}}(T)
    = \frac{C_{\text{die}} + C_{\text{VRAM}}}{T}
    + \frac{C_{\text{periph}}}{N_{\text{GPU}} \times T},
\end{equation}
where $N_{\text{GPU}}$ is the number of GPUs sharing the same server
peripherals. We report the annualized values for deployment durations of
$T \in \{3, 5, 8\}$ years in \Cref{tab:cpa-table}.

For the B200 GPU, since manufacturing data are not publicly available, we
estimate its CPA using the closest available process node. These deduced results
are marked with an~$\ast$ in \Cref{tab:cpa-table}.

\paragraph{Reliability overhead for second-hand devices.}
For second-hand devices such as the V100, an additional factor must be
considered: hardware reliability. During the reuse process, we apply a binning
procedure that involves stress-testing all components and removing those that
are faulty on arrival. An additional round of production testing is conducted to
filter out GPUs prone to early failure.

After binning, we deploy $10\%$ additional devices as spare capacity to cover
in-service failures, based on the $\sim\!90\%$ survival rate observed over a
5-year service lifespan reported by
Ostrouchov~\textit{et al.}~\cite{ostrouchov2020gpu}. For V100s previously
deployed for an initial 3-year data-center cycle, this extends their total
product lifecycle to 8~years ($3 + 5$).

The reliability-adjusted annualized embodied carbon for second-hand GPUs is
therefore:
\begin{equation}\label{eq:v100-reliability}
  C_{em}^{\text{annual,adj}}(T) = C_{em}^{\text{annual}}(T) \times 1.1,
\end{equation}
where the factor $1.1$ accounts for the $10\%$ additional devices deployed. The
resulting 3, 5, and 8-year embodied carbon costs for V100s are shown in
\Cref{tab:cpa-table}.

\paragraph{Carbon footprint perspectives for second-hand devices.}
When reporting the per-token carbon footprint ($C_{\text{total}}$,
\Cref{eq:ctotal}) in \Cref{tab:cost-8b:prefill,tab:cost-70b:prefill}, we present two
accounting perspectives for second-hand V100 devices:

\begin{enumerate}
  \item \textbf{Extended perspective.} The hardware lifecycle is viewed as
  extended from the original first-life deployment into a second-life reuse
  period. The full embodied carbon is amortized over the second-life deployment
  duration $T$ using the reliability-adjusted annualized cost
  (\Cref{eq:v100-reliability}):
  \begin{equation}\label{eq:extended}
    C_{\text{total}}^{\text{extended}}
      = C_{\text{opt}} + C_{\text{em}}^{\text{annual,adj}}(T).
  \end{equation}

  \item \textbf{Zero embodied perspective.} The V100s were scheduled for
  decommission and would otherwise be discarded (e.g.\ landfilled). Since their
  embodied carbon was already ``spent'' during the initial data-center cycle,
  reuse incurs no additional manufacturing emissions. The carbon footprint thus
  comprises only operational carbon:
  \begin{equation}\label{eq:zero-embodied}
    C_{\text{total}}^{\text{zero}} = C_{\text{opt}}.
  \end{equation}
\end{enumerate}

\noindent For new hardware (B200, H100, A100), only a single carbon footprint
is reported, computed using the standard annualized embodied carbon
$C_{\text{em}}^{\text{annual}}(T)$ from \Cref{eq:annual}. The two
perspectives above represent an \textbf{upper bound} (extended) and a
\textbf{lower bound} (zero embodied) on the attributable embodied carbon for
reused devices; the appropriate allocation depends on the carbon accounting
framework adopted.

It is worth noting that this lifecycle analysis excludes emissions from
end-of-life disposal (e.g.\ metal recycling and landfill), as comprehensive GPU
lifecycle data remain limited in the literature. Our CPA estimates thus
primarily account for manufacturing carbon and do not explicitly isolate upstream
material-sourcing emissions.

\end{appendices}

\clearpage
\newpage

\bibliographystyle{unsrtnat}
\bibliography{references}

@misc{AMD,
    title         = {{AMD} {I}nstinct {MI}300 {S}eries {A}ccelerators},
    note          = {Accessed: 2024-03-03},
    howpublished  = {\url{https://www.amd.com/en/products/accelerators/instinct/mi300.html}},
}

@article{choquette2023nvidia,
    title         = {{NVIDIA} {H}opper {H}100 {GPU}: {S}caling {P}erformance},
    author        = {Choquette, Jack},
    year          = 2023,
    journal       = {IEEE Micro},
    number        = 3,
    pages         = {9--17},
}

@article{dubey2024llama,
  title={The llama 3 herd of models},
  author={Dubey, Abhimanyu and Jauhri, Abhinav and Pandey, Abhinav and Kadian, Abhishek and Al-Dahle, Ahmad and Letman, Aiesha and Mathur, Akhil and Schelten, Alan and Yang, Amy and Fan, Angela and others},
  journal={arXiv preprint arXiv:2407.21783},
  year={2024}
}

@article{achiam2023gpt,
  title={Gpt-4 technical report},
  author={Achiam, Josh and Adler, Steven and Agarwal, Sandhini and Ahmad, Lama and Akkaya, Ilge and Aleman, Florencia Leoni and Almeida, Diogo and Altenschmidt, Janko and Altman, Sam and Anadkat, Shyamal and others},
  journal={arXiv preprint arXiv:2303.08774},
  year={2023}
}

@article{faiz2023llmcarbon,
  title={Llmcarbon: Modeling the end-to-end carbon footprint of large language models},
  author={Faiz, Ahmad and Kaneda, Sotaro and Wang, Ruhan and Osi, Rita and Sharma, Prateek and Chen, Fan and Jiang, Lei},
  journal={arXiv preprint arXiv:2309.14393},
  year={2023}
}

@incollection{bhagavathula2024understanding,
  title={Understanding the Implications of Uncertainty in Embodied Carbon Models for Sustainable Computing},
  author={Bhagavathula, Anvita and Han, Leo and Gupta, Udit},
  booktitle={HotCarbon},
  year={2024}
}

@inproceedings{li2024towards,
  title={Towards Carbon-efficient LLM Life Cycle},
  author={Li, Yueying Lisa and Graif, Omer and Gupta, Udit},
  booktitle={Proceedings of the 3rd Workshop on Sustainable Computer Systems (HotCarbon)},
  year={2024}
}

@inproceedings{kwon2023efficient,
  title={Efficient memory management for large language model serving with pagedattention},
  author={Kwon, Woosuk and Li, Zhuohan and Zhuang, Siyuan and Sheng, Ying and Zheng, Lianmin and Yu, Cody Hao and Gonzalez, Joseph and Zhang, Hao and Stoica, Ion},
  booktitle={Proceedings of the 29th Symposium on Operating Systems Principles},
  pages={611--626},
  year={2023}
}

@article{zheng2024sglang,
  title={Sglang: Efficient execution of structured language model programs},
  author={Zheng, Lianmin and Yin, Liangsheng and Xie, Zhiqiang and Sun, Chuyue Livia and Huang, Jeff and Yu, Cody Hao and Cao, Shiyi and Kozyrakis, Christos and Stoica, Ion and Gonzalez, Joseph E and others},
  journal={Advances in Neural Information Processing Systems},
  volume={37},
  pages={62557--62583},
  year={2024}
}

@article{wang2024waste,
  title={E-waste challenges of generative artificial intelligence},
  author={Wang, Peng and Zhang, Ling-Yu and Tzachor, Asaf and Chen, Wei-Qiang},
  journal={Nature Computational Science},
  pages={1--6},
  year={2024},
  publisher={Nature Publishing Group US New York}
}

@article{tomlinson2024carbon,
  title={The carbon emissions of writing and illustrating are lower for AI than for humans},
  author={Tomlinson, Bill and Black, Rebecca W and Patterson, Donald J and Torrance, Andrew W},
  journal={Scientific Reports},
  volume={14},
  number={1},
  pages={3732},
  year={2024},
  publisher={Nature Publishing Group UK London}
}

@inproceedings{li2023toward,
  title={Toward sustainable hpc: Carbon footprint estimation and environmental implications of hpc systems},
  author={Li, Baolin and Basu Roy, Rohan and Wang, Daniel and Samsi, Siddharth and Gadepally, Vijay and Tiwari, Devesh},
  booktitle={Proceedings of the international conference for high performance computing, networking, storage and analysis},
  pages={1--15},
  year={2023}
}

@misc{eadline_2024, title={Nvidia Shipped 3.76 Million Data-center GPUs in 2023, According to Study}, url={https://www.hpcwire.com/2024/06/10/nvidia-shipped-3-76-million-data-center-gpus-in-2023-according-to-study/}, journal={HPCwire}, author={Eadline, Doug}, year={2024}, month={Jun} }

@misc{jade_hpc_2024, title={JADE Service End of Life Announcement}, url={https://www.jade.ac.uk/}, journal={Jade.ac.uk}, author={Jade HPC}, year={2024} }

@misc{microsoft_2025, title={Migration Guide for GPU Compute Workloads in Azure - Azure Virtual Machines}, url={https://learn.microsoft.com/en-us/azure/virtual-machines/migration/sizes/n-series-migration}, journal={Microsoft.com}, author={microsoft}, year={2025}, month={Mar} }

@misc{microsoft_azure_2025, title={NCv3 and NC24rs Retirement - Azure Virtual Machines}, url={https://learn.microsoft.com/en-us/azure/virtual-machines/ncv3-nc24rs-retirement}, journal={Microsoft.com}, author={microsoft azure}, year={2025}, month={Mar} }

@misc{procurri_2024, title={Decommissioned the Largest NVIDIA DGX A100 Environment outside of a Hyperscaler}, url={https://www.linkedin.com/posts/procurri_weve-just-decommissioned-the-largest-nvidia-activity-7275473830779703297-cQuQ}, journal={Linkedin.com}, author={Procurri}, year={2024}, month={Dec} }

@article{hewage2025aging,
  title={Aging-aware CPU Core Management for Embodied Carbon Amortization in Cloud LLM Inference},
  author={Hewage, Tharindu B and Ilager, Shashikant and Read, Maria Rodriguez and Buyya, Rajkumar},
  journal={arXiv preprint arXiv:2501.15829},
  year={2025}
}

@inproceedings{ostrouchov2020gpu,
  title={GPU lifetimes on Titan supercomputer: Survival analysis and reliability},
  author={Ostrouchov, George and Maxwell, Don and Ashraf, Rizwan A and Engelmann, Christian and Shankar, Mallikarjun and Rogers, James H},
  booktitle={SC20: International Conference for High Performance Computing, Networking, Storage and Analysis},
  pages={1--14},
  year={2020},
  organization={IEEE}
}

@article{kokolis2024revisiting,
  title={Revisiting Reliability in Large-Scale Machine Learning Research Clusters},
  author={Kokolis, Apostolos and Kuchnik, Michael and Hoffman, John and Kumar, Adithya and Malani, Parth and Ma, Faye and DeVito, Zachary and Sengupta, Shubho and Saladi, Kalyan and Wu, Carole-Jean},
  journal={arXiv preprint arXiv:2410.21680},
  year={2024}
}

@inproceedings{heydari2022power,
  title={Power usage effectiveness analysis of a high-density air-liquid hybrid cooled data center},
  author={Heydari, Ali and Eslami, Bahareh and Radmard, Vahideh and Rebarber, Fred and Buell, Tyler and Gray, Kevin and Sather, Sam and Rodriguez, Jeremy},
  booktitle={International Electronic Packaging Technical Conference and Exhibition},
  volume={86557},
  pages={V001T01A014},
  year={2022},
  organization={American Society of Mechanical Engineers}
}

@inproceedings{patel2024splitwise,
  title={Splitwise: Efficient generative llm inference using phase splitting},
  author={Patel, Pratyush and Choukse, Esha and Zhang, Chaojie and Shah, Aashaka and Goiri, {\'I}{\~n}igo and Maleki, Saeed and Bianchini, Ricardo},
  booktitle={2024 ACM/IEEE 51st Annual International Symposium on Computer Architecture (ISCA)},
  pages={118--132},
  year={2024},
  organization={IEEE}
}

@techreport{nvidia_volta_wp2017,
  title        = {NVIDIA Tesla V100 GPU Architecture Whitepaper},
  author       = {{NVIDIA Corporation}},
  institution  = {NVIDIA GPU Technology Conference},
  year         = {2017},
  note         = {Die Size: 815\,mm\textsuperscript{2}; Accessed: 2025-05-17}
}

@techreport{nvidia_ampere_wp2020,
  title        = {NVIDIA A100 Tensor Core GPU Architecture Whitepaper},
  author       = {{NVIDIA Corporation}},
  institution  = {NVIDIA GPU Technology Conference},
  year         = {2020},
  note         = {Die Size: 826\,mm\textsuperscript{2}; Accessed: 2025-05-17}
}

@techreport{nvidia_hopper_wp2022,
  title        = {NVIDIA H100 Tensor Core GPU Architecture Whitepaper},
  author       = {{NVIDIA Corporation}},
  institution  = {NVIDIA GPU Technology Conference},
  year         = {2022},
  note         = {Die Size: 814\,mm\textsuperscript{2}; Accessed: 2025-05-17}
}

@article{owid-energy-mix,
    author = {Hannah Ritchie and Pablo Rosado},
    title = {Energy Mix},
    journal = {Our World in Data},
    year = {2020},
    note = {https://ourworldindata.org/energy-mix}
}

@misc{owid_co2_per_unit_energy_gcb_2025,
  title        = {{Carbon} intensity of energy production – GCB},
  author       = {{Global Carbon Project and U.S. Energy Information Administration and Energy Institute, processed by Our World in Data}},
  year         = {2025},
  howpublished = {\url{https://ourworldindata.org/grapher/co2-per-unit-energy}},
  note         = {Data sources: Global Carbon Budget (2024); U.S. Energy Information Administration “International Energy Data” (2023); Energy Institute “Statistical Review of World Energy” (2024); retrieved 19 May 2025}
}

@techreport{uk_parliament_post_2006,
  title        = {Carbon footprint of electricity generation},
  author       = {{Parliamentary Office of Science and Technology}},
  institution  = {UK Parliament},
  number       = {POSTnote\,268},
  month        = oct,
  year         = {2006},
  url          = {https://www.parliament.uk/globalassets/documents/post/postpn268.pdf},
  note         = {Offshore wind life-cycle emissions: 5.25 gCO2 eq/kWh; Accessed: 2025-05-19}
}

@misc{ceic_china_industrial_price_2025,
  title        = {China Electricity Price: Industrial Usage, 35 kV \& Above (Jan 2003–Jan 2025)},
  author       = {{CEIC Data}},
  year         = {2025},
  howpublished = {\url{https://www.ceicdata.com/en/china/electricity-price-36-city}},
  note         = {Industrial electricity usage price by city; Accessed: 2025-05-19}
}

@misc{eia_industrial_price_2025,
  title        = {Average Price of Electricity to Ultimate Customers by End-Use Sector, Table 5.6.A},
  author       = {{U.S. Energy Information Administration}},
  year         = {2025},
  howpublished = {\url{https://www.eia.gov/electricity/monthly/epm_table_grapher.php?t=epmt_5_6_a}},
  note         = {Industrial sector prices (cents per kWh) for March 2025; Accessed: 2025-05-19}
}

@misc{ember2026carbon,
  author       = {{Ember} and {Our World in Data}},
  title        = {Lifecycle carbon intensity of electricity generation -- Ember},
  year         = {2026},
  howpublished = {\url{https://archive.ourworldindata.org/20260304-094028/grapher/carbon-intensity-electricity.html}},
  note         = {Dataset. Original data: Ember, ``Yearly Electricity Data Europe''; Ember, ``Yearly Electricity Data''. Major processing by Our World in Data. Retrieved March 16, 2026 (Archived on March 4, 2026).},
}

@misc{globalpetrolprices_brazil,
  author       = {{GlobalPetrolPrices}},
  title        = {Brazil electricity prices},
  year         = {2026},
  howpublished = {\url{https://www.globalpetrolprices.com/Brazil/electricity_prices/}},
  note         = {Accessed: 2026-03-17}
}
\end{document}